\documentclass[nonblindrev]{informs3} 

\DoubleSpacedXI 

\usepackage{endnotes}
\let\footnote=\endnote

\usepackage{natbib}
 \bibpunct[, ]{(}{)}{,}{a}{}{,}%
 \def\bibfont{\small}%
\usepackage{subfigure}
\usepackage{microtype}
\usepackage{color}
\usepackage{appendix}
\usepackage{multirow}
\usepackage{booktabs}

\usepackage{algorithm}
\usepackage{algpseudocode}
\usepackage{tikz}

\usepackage{amsmath}
\usepackage{array}
\usepackage{bm}
\usepackage{lipsum}
\usepackage{float}
\usepackage{arydshln}
\usepackage{overpic}

\TheoremsNumberedThrough     
\ECRepeatTheorems

\EquationsNumberedThrough    

\begin{document}


\RUNAUTHOR{Wang et al.}

\RUNTITLE{DQL}


\TITLE{Kernel-Based Distributed Q-Learning: A Scalable
Reinforcement Learning Approach for Dynamic  Treatment Regimes}

\ARTICLEAUTHORS{
\AUTHOR{Di Wang,  Yao Wang,  Shao-Bo Lin \footnotemark[1]}
\AFF{ Center for Intelligent Decision-Making and Machine Learning, School of Management, Xi'an Jiaotong University, Xi'an, China}
\footnotetext[1]{Corresponding authors: sblin1983@gmail.com}
} 

\ABSTRACT{In recent years, large amounts of electronic health records (EHRs) concerning chronic diseases have been collected to facilitate medical diagnosis. Modeling the dynamic properties of EHRs related to chronic diseases can be efficiently done using dynamic treatment regimes (DTRs). While reinforcement learning (RL) is a widely used method for creating DTRs, there is ongoing research in developing RL algorithms that can effectively handle large amounts of data. In this paper, we present a scalable kernel-based distributed Q-learning algorithm  for generating DTRs. We perform both theoretical assessments and numerical analysis for the proposed approach. The results demonstrate that our algorithm significantly reduces the computational complexity associated with the state-of-the-art deep reinforcement learning methods, while maintaining comparable generalization performance in terms of accumulated rewards across stages, such as survival time or cumulative survival probability.
}%


\KEYWORDS{Q-learning, dynamic  treatment regimes, distributed learning, kernel ridge regression}

\maketitle


%


\section{Introduction}

Patients with chronic diseases, such as cancer, diabetes, and mental disease, usually undergo a period of initial treatment, disease recurrence, and salvage treatment. Explicitly specifying the relationship of treatment type and drug dosage with patient response is generally difficult, so  practitioners have to use protocols in which each patient receives a similar treatment based on the average responses of previous patients with similar cases.  However, illnesses respond heterogeneously to treatment.
For example, a study on schizophrenia \citep{Ishigooka2000} discovered that patients using the exact same antipsychotic experience drastically varied outcomes; some patients experience few adverse events with improved clinical outcomes, whereas others have to discontinue treatment due to worsening symptoms. This has motivated researchers to advocate dynamic  treatment regimes (DTRs) to determine treatments dynamically and individually based on clinical observations of patients \citep{murphy2005experimental,Chakraborty2013,Almirall2016}.

A DTR is defined by a set of sequential decision rules; each rule takes the clinical observations of a patient at certain key points as input and returns the treatment action of doctors as output \citep{tsiatis2019dynamic}. In this way, the design of a DTR can be considered a sequential decision-making problem and suits the reinforcement learning  (RL) framework \citep{sutton2018reinforcement}, in which states, actions, rewards, and policies correspond to the clinical observations of patients, treatment options, treatment outcomes, and series of decision rules, respectively. RL simultaneously addresses sequential decision-making problems with sampled, evaluative, and delayed feedback, which naturally makes it a desirable tool for developing ideal DTRs.

Q-learning \citep{watkins1992q} is a typical temporal-difference algorithm that produces a sequence of high-quality actions in RL and has  {been} substantially developed in the past decades in health care \citep{oh2022generalization}.
{Different from other applications, Q-learning for DTRs with a finite horizon (typically small and not exceeding a few tens) commonly has the following important characteristics:}

$\bullet$ Data structure:
When a patient's chronic disease is not effectively treated or managed through a particular treatment plan, they may seek alternative treatment options from other doctors, resulting in treatment records with limited duration. Additionally, in Q-learning for DTRs, doctors typically have a limited number of treatment options to choose from based on patients' treatment histories \citep[Chap.1.4]{tsiatis2019dynamic}. However, for a given treatment decision, the clinical observations of patients can be unpredictable, resulting in an infinite number of states. This makes Q-learning for DTRs distinct from classical tabular-based Q-learning, which focuses on finite actions and states \citep{gosavi2009reinforcement}. To summarize, DTRs have finite horizons, and each stage of Q-learning for DTRs has a finite number of actions but an infinite number of states.

$\bullet$ Task orientation:
{For DTRs with a small number of stages, like non-small-cell lung cancer (NSCLC) patients typically undergoing one to three treatment lines \citep{Socinski2007NSCLC}, treatment decisions at each critical point in Q-learning are crucial for determining patient outcomes, regardless of timing. Hence, assuming one treatment phase is more critical than others is inappropriate, and the standard Q-learning reward discounting method is unsuitable for such DTRs.}

$\bullet$ Quality guarantee: Quality is crucial in medical treatment, as even a small mistake  {may  result in extremely} negative outcomes for patients. This necessitates not only a strong understanding of   {treatment decisions  but also   solid theoretical guarantees supporting the effectiveness}. This requirement eliminates the use of some heuristic yet intricate designs of Q-learning \citep{oroojlooyjadid2022deep} for generating DTRs.

$\bullet$ Computational costs:
This increase in data size poses significant scalability challenges for the Q-learning algorithms, in terms of storage and computational complexity.  {This scalability requirement excludes the popular deep learning in DTR since it inherently entails a substantial number of parameters and hyperparameter, markedly escalating the computational resource requirements.}


The characteristics discussed above place strict limitations on the use of Q-learning for generating DTRs. The need for quality guarantees excludes the commonly used linear Q-learning \citep{murphy2005generalization}, and the demands for computational efficiency and interpretability make the widely adopted deep Q-learning \citep{oroojlooyjadid2022deep} unsuitable for DTRs. In this paper, we aim to create a scalable Q-learning algorithm that has solid theoretical guarantees for DTRs.

\subsection{Mathematical formulation of Q-learning for DTRs}
DTRs often involve $T$-stage treatment decision problems. Let $s_t\in\mathcal S_t$ be the pretreatment information  {(or status)} and $a_t\in\mathcal A_t$ be the treatment decision  {(or action)} at stage $t$, where $\mathcal S_t$ and $\mathcal A_t$ are sets of pretreatment information and treatment decisions, respectively. A history of personalized treatment is generated in the form of
$\mathcal T_{T}=\{s_{1:T+1},a_{1:T}\}$, where $s_{T+1}$ is the status after all treatment decisions,   {$s_{1:t}$ and $a_{1:t}$ denote all historical status and actions recorded from stage 1 to stage $t$, i.e., $s_{1:t}=\{s_1,s_2,\cdots,s_t\}$, $a_{1:t}=\{a_1,a_2,\cdots,a_t\}$. Similarly, $\mathcal S_{1:t}$ and $\mathcal A_{1:t}$ represent the status space and action space, respectively, formed from stage 1 to stage $t$, i.e., $\mathcal S_{1:t} = \mathcal S_1 \times \mathcal S_2 \times \cdots \times \mathcal S_t$ and $\mathcal A_{1:t} = \mathcal A_1 \times \mathcal A_2 \times \cdots \times \mathcal A_t$.} Let $R_t: (\mathcal S_{1:t+1},\mathcal A_{1:t}) \rightarrow \mathbb R$ be the clinical outcome following the decision $a_t$, which depends on the pretreatment information $s_{1:t+1}$ and treatment history $a_{1:t}$.
In cancer treatment, for example, $R_t$ can be set as the tumor size or the survival time when the $t$-th treatment decision is made.
Throughout the paper, we use capital letters ($A_1$, $A_2$, etc.) to denote random variables and lowercase letters ($a_1$, $a_2$, etc.) to represent their realization.

A DTR $\pi=(\pi_1,\dots,\pi_T)$,  {where $\pi_t:\mathcal S_{1:t}\times\mathcal A_{1:t-1}\rightarrow \mathcal A_t$,} is a set of rules for personalized treatment decisions at all $T$ stages. The optimal DTR is that which maximizes the overall outcome of interest $\sum_{t=1}^TR_t(s_{1:t+1},a_{1:t})$ in terms of designing suitable treatment rules  {$\pi=(\pi_1,\dots,\pi_T)$}. Our analysis is cast in a random setting to quantify the overall outcome    \citep{murphy2005generalization}. Denote by $\rho_t(s_t|s_{1:t-1}, a_{1:t-1})$ the transition probability of $s_t$ conditioned on $s_{1:t-1}$ and $ a_{1:t-1}$. The value of a DTR $\pi$ at stage $t$ is  defined by
$$
     V_{\pi,t}(s_{1:t},  a_{1:t-1})=E_\pi\left[\sum_{j=t}^TR_j(  S_{1:j+1},  A_{1:j})\big|S_{1:t}=  s_{1:t},
      A_{1:t-1}= a_{1:t-1}\right],
$$
where $E_\pi$ is the expectation with respect to the distribution
$$
     P_\pi=\rho_1(s_1)1_{a_1=\pi_1(s_1)}\prod_{t=2}^{T}\rho_t(s_t|
     s_{1:t-1},  a_{1:t-1})1_{a_t=\pi( s_{1:t},
     a_{1:t-1})}\rho_{T+1}(s_{T+1}| s_{1:T},  a_{1:T}),
$$
and $1_W$ is the indicator of the event $W$. Denote the optimal value function of $\pi$ at stage $t$ as
$$
     V^*_{t}( s_{1:t},  a_{1:t-1})=\max_{\pi\in\Pi}V_{\pi,t}( s_{1:t}, a_{1:t-1}),
$$
where $\Pi$ is the collection of all treatments. We aim to find a policy $\hat{\pi}$ to minimize $V_1^*(s_1)-V_{\hat{\pi},1}(s_1)$.

Let $p_t(a_t|  s_{1:t}, a_{1:t-1})$ be the probability that a decision $a_t$ is made given the history $\{s_{1:t},  a_{1:t-1}\}$.
Define the optimal time-dependent $Q$-function by
\begin{equation}\label{Qfunction}
     Q_t^*( s_{1:t}, a_{1:t})=E[R_t(S_{1:t+1},  A_{1:t})+V_{t+1}^*(
     S_{1:t+1}, A_{1:t})| S_{1:t}= s_{1:t},
     A_{1:t}=  a_{1:t}],
\end{equation}
where $E$ is the  {expectation with respect to the distribution $P:=P_{T+1}$ and
\begin{equation}\label{likelihood under P}
   P_t=\rho_1(s_1)p_1(a_1|s_1)
   \prod_{j=2}^{t}\rho_j(s_j|
     s_{1:j-1},  a_{1:j-1})p_j(a_j| s_{1:j},
     a_{1:j-1}).
\end{equation}}
Given the definition of  $V^*_{t}$, it is easy to  derive \cite{murphy2005generalization}
$$
     V^*_{t}(s_{1:t}, a_{1:t-1})=V_{\pi^*,t}(s_{1:t}, a_{1:t-1})
     =
     E_{\pi^*}\left[\sum_{j=t}^TR_j(S_{1:j+1},  A_{1:j})\big|  S_{1:t}=  s_{1:t},
      A_{1:t-1}= a_{1:t-1}\right],
$$
 {where $\pi^*$ is the optimal policy }
and consequently
\begin{equation}\label{QandV}
    V^*_{t}( s_{1:t},  a_{1:t-1})=\max_{a_t}Q_t^*( s_{1:t}, a_{1:t}),
\end{equation}
 {showing that the optimal treatment decisions can be determined by maximizing the optimal $Q$-functions}. Thus, researchers have developed efficient algorithms for finding optimal $Q$-functions in the realm of RL \citep{murphy2005generalization}.
With $ Q_{T+1}^*=0$, the definition of $Q_t^*$ indicates that
\begin{equation}\label{conditonal expectation}
     Q_t^*( s_{1:t},  a_{1:t})
     =
     E\left[R_t( S_{1:t+1},  A_{1:t})+\max_{a_{t+1}}Q_{t+1}^*(  S_{1:t+1},
     A_{1:t},a_{t+1})\big|  S_{1:t}=  s_{1:t},  A_{1:t}=
     a_{1:t}\right]
\end{equation}
for $t=T,T-1,\dots,1$. This property connects $Q$-functions with the well-known regression function \citep{gyorfi2006distribution} in supervised learning as follows. Define $\mathcal X_t=\mathcal S_{1:t}\times\mathcal A_{1:t}$, and denote $\mathcal L_t^2$ as the space of square-integrable functions defined on $\mathcal X_t$ with respect to  {$P_t$}.
\begin{figure}[t]
    \FIGURE
    {\includegraphics[width=16cm,height=7cm]{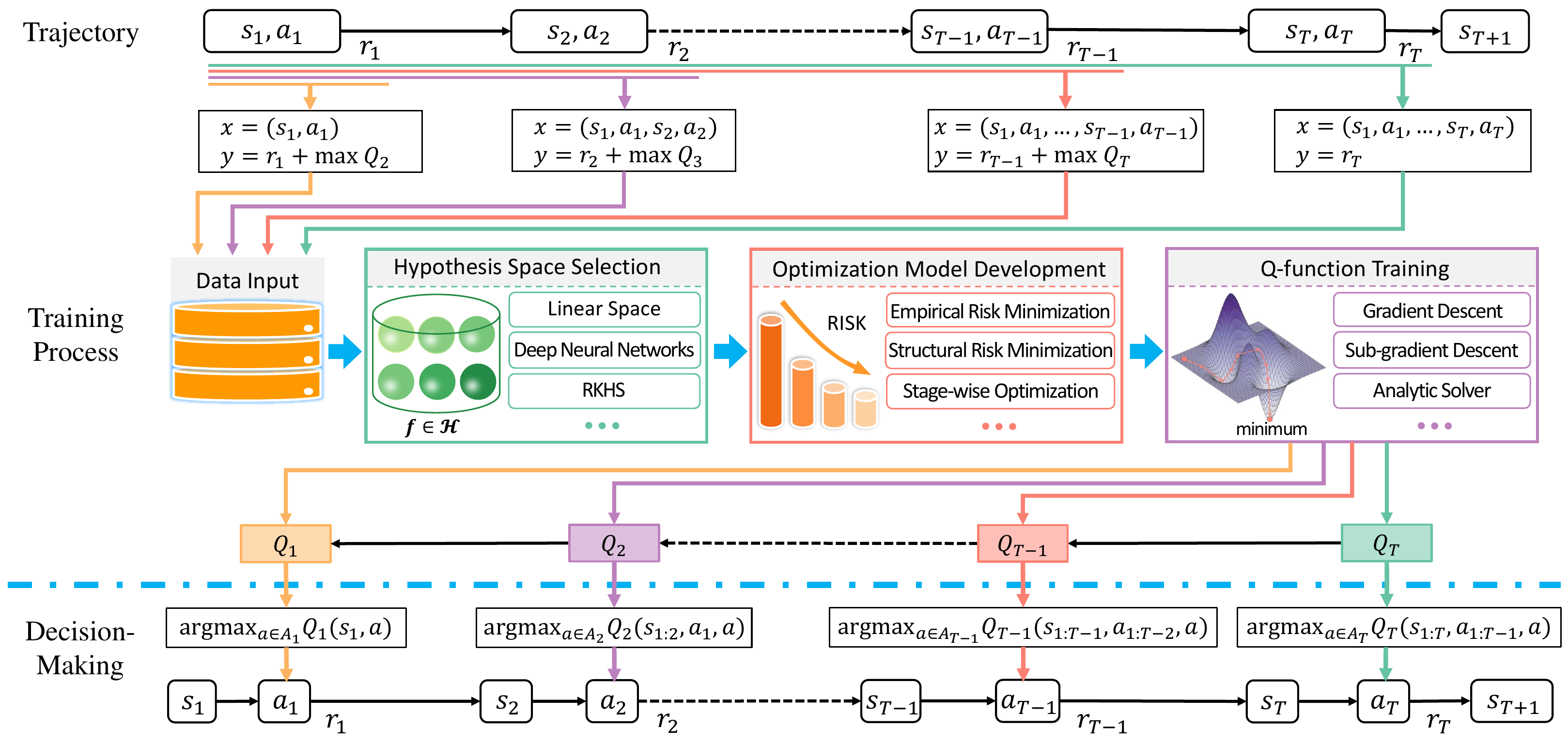}}
    {Training and decision-making flows of Q-learning. \label{Fig:scheme_Qlearning}}
    {}
\end{figure}

Write
\begin{equation}\label{def.output}
  x_t:=\{s_{1:t},a_{1:t}\}\in\mathcal X_t,\qquad\mbox{and}\quad  y_t^*:=R_t( s_{1:t+1},  a_{1:t})+\max_{a_{t+1}}Q_{t+1}^*(  s_{1:t+1},
     a_{1:t},a_{t+1}).
\end{equation}
$Q_t^*$ can be regarded as the regression function of data $(x_t,y_t^*)$ and written as
\begin{equation}\label{Regression-function-Q}
    Q_t^*=E[Y_t^*|X_t],\qquad t=T,T-1,\dots,1.
\end{equation}
Therefore, the standard approach in statistical learning theory
\citep{gyorfi2006distribution} yields
\begin{equation}\label{population}
      Q_t^*={\arg\min}_{Q_t\in \mathcal L_t^2}E\left[\left(Y_t^*-Q_t(X_t)\right)^2\right],\qquad t=T,T-1,\dots,1,
\end{equation}
showing that optimal $Q$-functions can be obtained by solving $T$ least squares problems. As shown in Figure \ref{Fig:scheme_Qlearning}, through the iterative generation of input--output pairs, Q-learning can be regarded as a series of supervised learning problems and incorporates four steps:
hypothesis space selection, optimization model development, Q-function training, and policy search. The hypothesis space selection step focuses on selecting a class of functions that encodes some a priori information and determines the formats of Q-functions.
The optimization model development step involves mathematically defining the Q-functions via a sequence of optimization problems.
After the optimization model is developed, a feasible optimization algorithm
should be designed in the Q-function training step to derive Q-functions. Finally, the policy search step searches for the best possible policies for choosing appropriate actions by maximizing the Q-functions.



\subsection{Problem setting}
 {We} are interested in Q-learning with $T$ stages whose state space $\mathcal S_t$ is an infinite continuous space and whose action space $\mathcal A_t$ is a discrete set of finite elements.
To suit different chronic diseases, the proposed Q-learning algorithm for DTRs should be feasible, efficient, and theoretically sound for a large number of reward functions because different reward functions usually correspond to various chronic diseases \citep{padmanabhan2017reinforcement}. The setting of  Q-learning in this paper is shown in Table \ref{Tab:setting}.

\begin{table}[t]
\centering
\caption{Setting of  Q-learning for DTRs.}\label{Tab:setting}
 \begin{tabular}{| c | c | c | c | c | c | c |}
 \toprule[1.5pt]
 Horizon & Action space & State space & Reward function & Discount  & Quality\\
  \hline
T & Finite & Continuous & Smooth & No  & Generalization error\\
 \toprule[1.5pt]
 \end{tabular}
\end{table}
As the starting point of Q-learning, hypothesis spaces regulate the format and properties of the Q-functions to be learned and determine the type of optimization models to be developed.
Linear spaces \citep{murphy2005generalization} and deep neural networks \citep{franccois2018introduction} are widely known hypothesis spaces for Q-learning.
However, linear spaces have limited expressive power and cannot adapt to different reward functions, frequently performing poorly for chronic diseases, such as cancer and sepsis \citep[Sec.4]{yu2021reinforcement}, whereas deep neural networks lack solid theoretical guarantees and are usually time consuming.
 {Since Q-learning for DTRs requires   a perfect balance between an expressive hypothesis space for quality guarantees and a simple hypothesis space for scalability}, we are faced with the following challenge:
\begin{problem}\label{Prob.1}
How can a hypothesis space be designed to equip  Q-learning for DTRs to avoid  the limitations of both linear spaces and deep neural networks simultaneously?
\end{problem}

Once the hypothesis space is determined, optimization models  {are crucial} in defining Q-functions. In general, a good optimization model for Q-learning for DTRs should be solvable, scalable, and interpretable. Solvability means that the developed model can be solved using standard optimization algorithms like gradient descent. Scalability implies that the cost of solving the optimization problem is reasonable. Interpretability means that the developed model is easy to understand. Optimization models with good solvability, scalability, and interpretability can be easily developed for linear hypothesis spaces, but for nonlinear spaces, these properties are difficult to achieve. Therefore, we focus on the following problem:

\begin{problem}\label{Prob-2}
How to develop solvable, scalable, and interpretable optimization models for Q-learning for DTRs  to optimize the performance of chosen nonlinear or infinite hypothesis spaces?
\end{problem}

 {The solutions to Problems \ref{Prob.1} and \ref{Prob-2} are sufficient to develop DTRs through Q-learning}. With the emphasis on the quality guarantee of Q-learning for DTRs, we ultimately focus on the following problem:

\begin{problem}\label{Prob.3}
How can the quality of the resulting DTRs be theoretically measured and how can solid theoretical guarantees be provided for it?
\end{problem}

Unlike other applications, such as recommendations and supply chain management, which use regret \citep{liu2022provably} to measure the quality of the developed policies, DTRs focus on the final treatment outcomes, regardless of the median treatment processes, to fully explore the values of the initial states. This means that new analysis tools are required for theoretical guarantees.

\subsection{Our contributions}
We develop a scalable Q-learning algorithm based on kernel ridge regression (KRR) \citep{caponnetto2007optimal} for DTRs to address the above problems.
Our main contributions are as follows:

$\bullet$ Methodological novelty:  {Breaking down Q-learning into $T$ least squares regression problems}, we combine  distributed learning \citep{zhang2015divide} and KRR to estimate Q-functions. Specifically, we adopt distributed regularized least squares over specific RKHS
as the optimization model in Q-learning. This approach avoids the quality limitations of linear Q-learning and addresses the computational efficiency issues of deep Q-learning, simultaneously.

$\bullet$ Theoretical Novelty: We propose a novel integral operator approach to analyze the generalization error of Q-learning.
Our theoretical results indicate that  distributed learning does not increase the generalization error, provided the data are not divided into too many subsets. Additionally, the generalization error bounds we derived are independent of the input dimensions, indicating that the proposed algorithm scales well with respect to the dimensionality of the problem.

$\bullet$ Numerical Novelty: We evaluate the proposed algorithm against linear Q-learning and  {several state-of-the-art deep RL methods} in two types of clinical trials for cancer treatments. Our numerical results show that the performance of our approach is consistently better than that of linear Q-learning and  {comparable to that  of deep RL methods}. In terms of computational costs, our approach requires significantly less training time than  {deep RL methods}. These results demonstrate the effectiveness and efficiency of the proposed kernel-based distributed Q-learning for DTRs.

The rest of this paper is organized as follows. Section \ref{Sec.related work} introduces some related work and compares our approach with theirs. In Section \ref{Sec.dkql}, we explain the proposed kernel-based distributed Q-learning algorithm for DTRs. In Section \ref{Sec.theory}, we study the theoretical behaviors of the proposed algorithm by establishing two generalization error bounds, whose proofs are given in the appendix.
In Section \ref{sec.simulation}, two simulations are conducted to illustrate how the proposed algorithm outperforms linear Q-learning and deep  {RL methods}.

\section{Related Work and Comparisons}\label{Sec.related work}

 {EHRs have been widely used in health care decisions regarding chronic diseases over the last decade.}
Q-learning is a promising approach to utilizing EHRs to yield high-quality DTRs and has gained fruitful clinical effects in improving the quality of health care decisions \citep{tsiatis2019dynamic,yu2021reinforcement}. In particular,
\cite{Zhao2009} used support vector regression (SVR) and extremely randomized trees to fit approximated Q-functions for agent dosage decision-making in chemotherapy; \cite{Humphrey2017thesis} employed classification and regression trees (CART), random forests, and a modified version of multivariate adaptive regression splines to estimate Q-values for an advanced generic cancer trial; \cite{Tseng2017} presented a multicomponent deep RL framework for patients with NSCLC in which a deep Q-network (DQN) maps states into potential dose schemes to optimize future radiotherapy outcomes; \cite{Raghu2017} proposed a fully connected dueling double DQN approach to learn an approximation of the optimal action--value function for sepsis treatment. The reader is referred to Tables III and IV of \citep{yu2021reinforcement} for more related work on Q-learning for DTRs.

 {Compared with these methods,} our approach possesses  three main novelties. First, we employ kernel-based Q-learning to generate DTRs because the design of kernel methods satisfies the high requirements of Q-learning for DTRs (the hypothesis space of a kernel method is large enough, and the derived estimator is in a linear space). Second, we focus on distributed Q-learning algorithms for DTRs that scale well according to the data size. Such a scalable variant of Q-learning for DTRs has not been considered in the literature. Finally, we provide solid theoretical guarantees for the proposed Q-learning approach in the framework of statistical learning theory \citep{murphy2005generalization}, which is beyond the scope of the above mentioned papers.

 {Note that there are also asynchronous parallel deep Q-learning algorithms, such as the distributed method proposed by \cite{ong2015distributeddeepqlearning}, which can accelerate computations through distributed asynchronous gradient calculations. However, our method differs significantly. Specifically, their Q-function remains consistent across different stages, whereas ours is stage-dependent. Regarding communication, their methods entail costs per parameter iteration, while ours involves only one communication per stage. More importantly, their distributed nodes require a real environment to collect experiences, but ours trains solely on historical data without a real environment model.}

 {We then introduce several theoretical results of Q-learning.}
From the theoretical analysis viewpoint, the quality of Q-learning has been extensively studied, and  can be roughly classified into two types \citep{levine2020offline,Prudencio2023}. The one is for online Q-learning, which constantly interacts with the environment to collect new experiences with the latest policy before updating the Q-function.
In particular, \citet{kearns1999finite} proposed a framework to deduce the
finite-time convergence rates (i.e., sample size bounds)  for Q-learning algorithms; \citet{even2003learning} derived a sufficient condition of the learning rate in gradient-based linear Q-learning to guarantee the convergence (stability) of Q-learning; \citet{xu2007kernel} proposed a kernel-based least squares policy iteration algorithm and presented the convergence analysis; \citet{wainwright2019variance} developed a variance reduction technique to improve the performance of Q-learning and proved that it achieves the minimax optimal sample complexity up to a logarithmic factor in the discount complexity; \citet{liu2022provably} studied regret bounds for kernel-based Q-learning.
The other is for offline Q-learning (fitted-Q-iteration, FQI), which learns Q-functions from a static data set of transitions and no longer interacts with the environment. Specifically, \citet{murphy2005generalization,Goldberg2012}, {\citet{wang2020statistical}}, and \citet{oh2022generalization} studied the generalization error of linear Q-learning; \citet{fan2020theoretical} deduced a generalization error estimate for deep Q-learning;
\citet{pmlrduan2021a} provided upper bounds on the generalization performance of Q-learning by using Rademacher complexity. {However, the setting of Q-learning studied in the current paper is different from most of these works, as we are concerned with the states, actions, and rewards in Table \ref{Tab:setting}. For example, except for \citep{murphy2005generalization,Goldberg2012,oh2022generalization}, all the above studies focused on Q-learning with large (or even infinite) horizons and discounted rewards.}




The most closely related works to ours are \citep{oh2022generalization} and \citep{liu2022provably}. The former deduced a similar generalization error to ours for linear Q-learning under the same setting as in the present paper, whereas the latter aimed to derive regret bounds for kernel-based Q-learning. The differences between our work and \citep{oh2022generalization} are as follows. We study kernel-based distributed Q-learning rather than linear Q-learning. The main advantage of utilizing an RKHS as the hypothesis space is the excellent expressive power over linear spaces. Moreover, we employ a novel integral operator approach to deduce the generalization error of Q-learning for DTRs rather than using the standard covering number approach. The generalization error bounds in our paper, derived under weaker assumptions on data distribution, are considerably tighter than the results in \citep{oh2022generalization}. The main differences between our work and \citep{liu2022provably} are three aspects. First, we are interested in generalization error analysis, whereas \citet{liu2022provably} focused on regret analysis.
Because the goal of treatment is to find a DTR $\hat{\pi}$ that minimizes $V_1^*(s_1)-V_{\hat{\pi},1}(s_1)$, the generalization error $E\left[V_1^*(s_1)-V_{\hat{\pi},1}(s_1)\right]$ is more appropriate for DTRs than the regret $\sum_{t=1}^T \left(\left(V_t^*(s_{1:t},a_{1:t-1})-V_{\hat{\pi},t}(s_{1:t},a_{1:t-1}\right)\right)$.
Furthermore, \citet{liu2022provably} imposed strong restrictions on the kernel (linear kernel or Gaussian kernel), whereas our approach is available to any positive-definite kernel. Finally, as we use a distributed version
of kernel methods, the computational costs of the proposed
approach for DTRs are much less than those of the kernel-based algorithms in \citep{liu2022provably}.

\section{Kernel-Based Distributed Q-Learning for DTRs}\label{Sec.dkql}

Since \eqref{population} is defined by expectation, $Q_t^*$ cannot be derived directly. The only thing we have access to is a set of records of personalized treatments $D:=\{\mathcal T_{i,T},r_{i,1:T}\}_{i=1}^{|D|}$, where samples in
$\{\mathcal T_{i,T}\}_{i=1}^{|D|}=\{(s_{i,1:T+1},a_{i,1:T})\}_{i=1}^{|D|}$  {are} assumed to be drawn independently and identically according to $P$, $r_{i,t}:=R_t(s_{i,1:t+1},a_{i,1:t})$, and $|D|$ denotes the cardinality of the set $D$.
We derive an approximation of $Q_t^*$ using $D$, which is a standard regression problem in supervised learning. Empirically, by setting $\hat{Q}_{T+1}=0$, we can compute $Q$-functions by solving  $T$ structural risk minimization problems
\begin{small}
\begin{equation}\label{Q learning algorithm for Q}
     \hat{Q}_t( s_{1:t},  a_{1:t})=
     \arg\min_{Q_t\in\mathcal H_t}\mathbb E_D\left[\left(R_t( S_{1:t+1},  A_{1:t})+\max_{a_{t+1}}\hat{Q}_{t+1}(S_{1:t+1},A_{1:t},a_{t+1})-Q_t( S_{1:t},  A_{1:t})\right)^2\right]+\lambda\Omega(Q_t)
\end{equation}
\end{small}
for $t=T,T-1,\dots,1$, where $\mathbb E_D$ is the empirical expectation, $\mathcal H_t$ is a parameterized hypothesis space, $\Omega(Q_t)$ is some structures of $Q_t$, and $\lambda$ is a tunable parameter to balance the empirical risk and structure. Our goal is to find a suitable $\mathcal H_t$ and $\Omega(Q_t)$ so that $\hat{Q}_t$ is close to ${Q}^*_t$.

Before presenting the detailed algorithm, we elucidate the reasons for using kernel-based Q-learning:

$\bullet$  Q-learning for DTRs can be regarded as a $T$-stage least squares regression problem, and kernel methods are well-developed learning schemes that exhibit optimal generalization performance \citep{caponnetto2007optimal,lin2017distributed} in least squares regression, which implies that  {kernel methods perform excellently with strong theoretical guarantees in each stage.}

$\bullet$  Numerous scalable variants of kernel methods \citep{zhang2015divide,rudi2015less,meister2016optimal} have been proposed and successfully used. These variants provide guidance for the development of scalable Q-learning algorithms for DTRs.

$\bullet$
Linear hypothesis spaces are beneficial for computations but have poor generalization performance. Conversely, the complex structures of deep neural networks enable deep Q-learning in generalization but are frequently time consuming. Kernel methods balance Q-learning with linear spaces and deep Q-learning in terms that kernelized Q-learning outperforms linear Q-learning in generalization and has a smaller computational burden than deep Q-learning.

Based on these, we use  KRR \citep{caponnetto2007optimal,zhang2015divide} to derive an approximation of $Q^*_t$ empirically based on the given dataset $D$.
Let $K(x,x'):=K_t(x,x')$ be a Mercer kernel, where $x,x'\in\mathcal X_t$, and $\mathcal H_{K,t}$ be the corresponding RKHS endowed with the norm $\|\cdot\|_{K,t}$. Given regularization parameters $\lambda_t$ for $t=1,\dots,T$, if we
set $Q_{D,\lambda_{T+1},T+1}=0$ with $\lambda_{T+1}=0$, then the $Q$-functions can be empirically defined by
\begin{equation}\label{KRR-Q}
     {Q}_{D,\lambda_t,t} ={\arg\min}_{Q_t\in\mathcal H_{K,t}}\frac1{|D|}\sum_{i=1}^{|D|}\left(y_{i,t}-Q_t(s_{i,1:t},
   a_{i,1:t})\right)^2+\lambda_t
     \|Q_t\|_{K,t}^2, \qquad t=T,T-1,\dots,1,
\end{equation}
where
\begin{equation}\label{def.y-empirical}
      y_{i,t}:=r_{i,t}+\max\limits_{a_{t+1}\in\mathcal{A}_{t+1}}Q_{D,\lambda_{t+1},t+1}(  s_{i,1:t+1},
     a_{i,1:t},a_{t+1}).
\end{equation}
  {Through straightforward computations,  ${Q}_{D,\lambda_t,t}$ can be explicitly formulated as
\begin{equation}\label{Qfucntion_solution}
{Q}_{D,\lambda_t,t} = \sum_{i=1}^{|D|} \alpha_{i,t}\cdot K((s_{i,1:t},a_{i,1:t}),\cdot)
\end{equation}
Here, $\bm{\alpha}_t:=\left[\alpha_{1,t},\cdots,\alpha_{|D|,t}\right]^\top=(\mathbb{K}_t+\lambda_t|D|\mathbb{I})^{-1}\bm{y}_t$ and
$K((s_{i,1:t},a_{i,1:t}),\cdot)$ represents the kernel function evaluated at the point $(s_{i,1:t}, a_{i,1:t})$, where $\bm{y}_t:=\left[y_{1,t},\cdots,y_{|D|,t}\right]^\top$ and $\mathbb{K}_t$ denotes the  kernel matrix whose $(i,j)$-th element is  $K((s_{i,1:t}, a_{i,1:t}), (s_{j,1:t}, a_{j,1:t}))$.}
In this way, \eqref{KRR-Q} can be solved backward to derive a sequence of estimators  {$\{Q_{D,\lambda_t,t}\}_{t=1}^T$}. We then
define the corresponding  {KRR for DTRs (KRR-DTR)} by $\pi_{D,\vec{\lambda}}=({\pi}_{D,\lambda_1,1},\dots,{\pi}_{D,\lambda_T,T})$ with
\begin{equation}\label{Q learning algorithm for policy}
   {\pi}_{D,\lambda_t,t}(s_{1:t},
   a_{1:t-1})={\arg\max}_{a_t\in {\mathcal A}_t}Q_{D,\lambda_t,t}(  s_{1:t},
   a_{1:t-1},a_t),\qquad  t=1,\dots,T.
\end{equation}

Solving KRR \eqref{KRR-Q} requires $\mathcal O(|D|^3)$ floating computations in each stage, which makes KRR time consuming for large-scale datasets.
A scalable variant of \eqref{KRR-Q} should be developed to reduce the computational burden so that \eqref{KRR-Q} can be successfully utilized to handle large-scale datasets to generate a  DTR of high-quality.
Distributed learning equipped with a divide-and-conquer strategy \citep{zhang2015divide,lin2017distributed,lin2020distributed} is preferable for this purpose. Kernel-based distributed Q-learning can be divided into four steps: division, local processing, synthesis, and iteration.

$\bullet$ {\it Division:} Randomly divide the dataset $D$ into $m$ disjoint subsets $D_j$ according to the uniform distribution, that is,  $D=\cup_{j=1}^mD_j$, $D_j\cap D_{j'}=\varnothing$ for $j\neq j'$. Denote $D_j=\{s_{i,j,1:T+1},a_{i,j,1:T},r_{i,j,1:T}\}_{i=1}^{|D_j|}$.

$\bullet$ {\it Local Processing: } Initialize $\overline{Q}_{D,{\lambda}_{T+1},T+1}\equiv 0$, where ${\lambda}_{T+1}= {0}$. At the $t$-th stage, calculate the output of the $i$-th sample in the $j$-th data subset as
\begin{equation}\label{DKRR-output}
     \overline{y}_{i,j,t}:=r_{i,j,t}+\max_{a_{t+1}\in\mathcal{A}_{t+1}}\overline{Q}_{D,{\lambda}_{t+1},t+1}(s_{i,j,1:t+1}, a_{i,j,1:t},a_{t+1}), \quad j=1,\cdots,m,
\end{equation}
and run KRR with parameter $\lambda_{t}$ on the data $D_{j}$ to obtain a local estimator of the Q-function
\begin{equation}\label{DKRR-Q-Local-T}
     {Q}_{D_{j},\lambda_{t},t} =\arg\min_{Q_t\in\mathcal H_{K,t}}\frac1{|D_j|}\sum_{i=1}^{|D_j|}\left(\overline{y}_{i,j,t}-Q_t(s_{i,j,1:t},a_{i,j,1:t})\right)^2+\lambda_{t}
     \|Q_t\|_{K,t}^2, \quad j=1,\cdots,m.
\end{equation}

$\bullet$ {\it Synthesis:} Synthesize the global estimator as
\begin{equation}\label{DKRR-estimator}
    \overline{Q}_{D,{\lambda}_t,t}=\sum_{j=1}^m\frac{|D_j|}{|D|}{Q}_{D_{j},\lambda_{t},t}.
\end{equation}

$\bullet$ {\it Iteration and Decision-Making:}  Repeat the local processing and synthesis steps for $t=T,T-1,\dots,1$, and obtain a set of Q-functions $\left\{\overline{Q}_{D, \lambda_t,t}\right\}_{t=1}^T$ based on DKRR. Define the  {DKRR for DTRs (DKRR-DTR)} by $\overline{\pi}_{D,\vec{\lambda}}=(\overline{\pi}_{D, {\lambda}_1,1}, \cdots,  \overline{\pi}_{D, {\lambda}_T,T})$, where
\begin{equation}\label{DKRR-DTR}
   \overline{\pi}_{D,{\lambda}_t,t}(s_{1:t},
   a_{1:t-1})={\arg\max}_{a_t\in {\mathcal A}_t}\overline{Q}_{D,{\lambda}_t,t}(  s_{1:t},
   a_{1:t-1},a_t),\qquad  t=1,\dots,T.
\end{equation}

\begin{figure}[t]
    \FIGURE
    {\includegraphics[width=16cm,height=7cm]{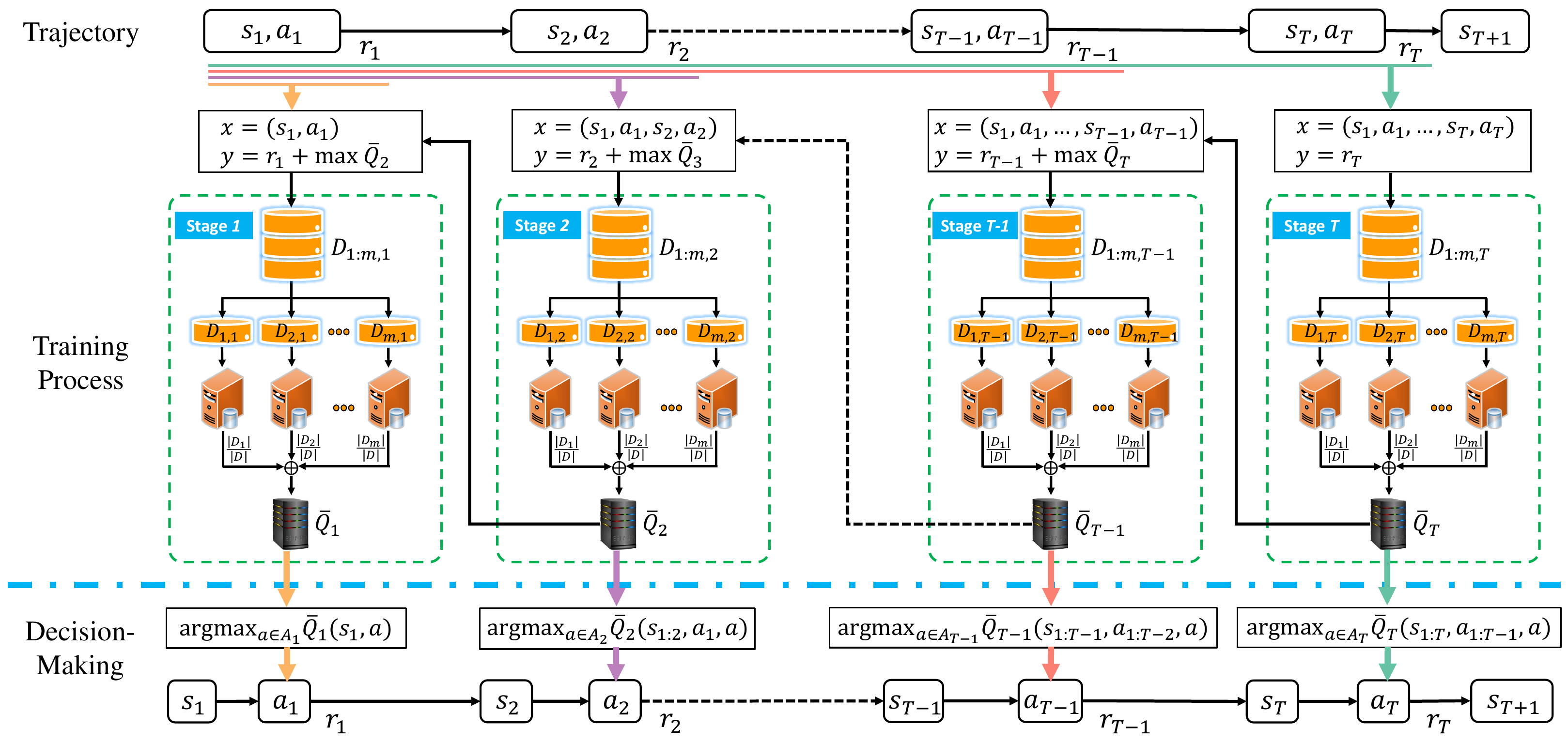}}
    {Training and decision-making flows of DKRR-DTR.\label{Fig:scheme}}
    {}
\end{figure}

The training and decision-making flows of the above approach are shown in Figure \ref{Fig:scheme}, where $D_{j,t}$ represents the data on the $j$-th local machine at the $t$-th stage and $\overline{Q}_{D, {\lambda}_t,t}$ is rewritten as $\overline{Q}_{t}$ for convenience.
A comparison of \eqref{DKRR-estimator} with \eqref{KRR-Q} shows that the computational complexity of KRR-DTR is reduced from $\mathcal O(|D|^3)$ to $\mathcal O(|D_{j}|^3)$ at each stage on the $j$-th local machine.
If the data subset sizes are the same, then each local machine has only $\mathcal O(|D|^3/m^3)$ floating computations at each stage.
As in the classical regression setting \citep{zhang2015divide,lin2017distributed}, the number of local machines $m$ balances the generalization error and computational complexity of DKRR-DTR. A small $m$ yields perfect generalization but needs huge computations, whereas a large $m$ significantly accelerates the algorithm but may degrade its generalization performance. Therefore, it is crucial to theoretically derive some values of $m$ with which DKRR-DTR performs similarly to KRR-DTR while reducing its computational burden.

\section{Theoretical Behaviors}\label{Sec.theory}

According to the ``no free lunch'' theory \citep[Theorem 3.1]{gyorfi2006distribution}, satisfactory generalization error bounds cannot be derived for learning algorithms without any assumptions about the data. Therefore, some restrictions on the data $D$ and the distribution $P$ should be set for our analysis. Our first assumption is mild under our setting.

\begin{assumption}\label{Assumption:compactness}
For any $t=1,\dots,T$,  $\mathcal S_{t}$ is a compact set, $\mathcal A_t$ is a finite set, and there exists an $M\geq 1$ such that  $\|R_t\|_{L^\infty}\leq M$ for any $t=1,\dots,T$.
\end{assumption}

Given Assumption \ref{Assumption:compactness}, it is easy to check that $\mathcal X_t=\mathcal S_{1:t}\times \mathcal A_{1:t}$ is also compact. According to (\ref{conditonal expectation}) and    $Q_{T+1}^*=0$, Assumption \ref{Assumption:compactness} implies that
\begin{equation}\label{bounded-y}
    |y_t^*|\leq (T-t+1)M, \qquad  \  \|Q^*_t\|_{L^\infty}\leq  (T-t+1)M
\end{equation}
hold almost surely for any $1\leq t\leq T$, showing that the optimal $Q$-functions are  bounded.  Our second assumption focuses on the conditional probability of choosing actions.

\begin{assumption}\label{Assumption:case-1}
Let $\mu\geq 1$ be a constant. It holds that
\begin{equation}\label{assumption-1}
   p_t(a|s_{1:t}, a_{1:t-1})\geq \mu^{-1},\qquad\forall \  a\in {\mathcal A}_t,\ t=1,\dots,T.
\end{equation}
\end{assumption}

Assumption \ref{Assumption:case-1}, which has been widely used
in RL \citep{murphy2005generalization,Goldberg2012}, declares that conditioned on previous treatment information, any action in the finite set $\mathcal A_t$ can be chosen with a probability of at least $\mu^{-1}$.  Based on Assumption \ref{Assumption:case-1}, it follows from \citep[Eq. (16)]{Goldberg2012} that for an arbitrary $Q_t\in \mathcal L_t^2$, there holds
\begin{equation}\label{Decomposition 1}
     E[V_1^*(S_1)-V_{ {\pi,1}}(S_1)]\leq
     \sum_{t=1}^T2\mu^{t/2}\sqrt{E[( {Q}_t-Q_t^*)^2]}=\sum_{t=1}^T2\mu^{t/2}\|{Q}_t-Q_t^*\|_{\mathcal L_t^2},
\end{equation}
where $\pi=(\pi_1,\dots,\pi_T)$ is defined by
$
   \pi_t( s_{1:t},
   a_{1:t-1})={\arg\max}_{a_t\in {\mathcal A}_t} {Q}_t(  s_{1:t},
   a_{1:t-1},a_t).
$


Our third assumption is the standard regularity assumption on $Q_t^*$ that has been adopted in \citep{caponnetto2007optimal,lin2017distributed}.
We first define an integral operator from $\mathcal H_{K,t}$ to $\mathcal H_{K,t}$ (also from $\mathcal L_t^2$ to $\mathcal L_t^2$ if no confusion is made) as
$
     L_{K,t}f(x)=\int_{\mathcal X_t}f(x')K(x,x')dP_t(x').
$
For any positive-definite kernel $K(\cdot,\cdot)$,
define $L_{K,t}^r$  as  the $r$-th power of $L_{K,t}$ via spectral calculus.

\begin{assumption}\label{Assumption:regularity}
For any $t=1,\dots,T$, assume that
\begin{equation}\label{regularitycondition}
         Q_t^*:=L_{K,t}^r h_t,~~{\rm for~some}  ~ h_t\in \mathcal L_{t}^2,~ r>0.
\end{equation}
\end{assumption}

The index $r$ in \eqref{regularitycondition} quantifies the regularity of $Q_t^*$. If $r=0$, then \eqref{regularitycondition} means $Q_t^*\in \mathcal L_t^2$, implying that there is no regularity of $Q_t^*$. If $r=1/2$, then \eqref{regularitycondition} is equivalent to $Q_t^*\in\mathcal H_{K,t}$, showing that $Q_t^*$ is in the RKHS associated with the kernel $K$.
If $r>1/2$, then \eqref{regularitycondition} implies that $Q_t^*$ is in other RKHSs  corresponding to smoother kernels than $K$. Generally speaking, the larger the $r$ value, the better the assumed regularity of $Q_t^*$. As indicated by \eqref{conditonal expectation}, the regularity assumption of $Q_t^*$ depends on the regularity of the reward function $R_t$. Since $R_t$ in a DTR can be manually set before the learning process, the regularity of $Q_t^*$ in Assumption \ref{Assumption:regularity} indeed includes a wide range of selection of $R_t$, including both the smooth and constant reward functions, and is much looser than the assumptions in existing literature \citep{murphy2005generalization,Goldberg2012,oh2022generalization}, where $Q_t^*$ is assumed to belong to linear spaces.

  For any $t$, define the effective dimension \citep{caponnetto2007optimal} as
$\mathcal{N}_t(\lambda):={\rm Tr}((\lambda I+L_{K,t})^{-1}L_{K,t})$ for $\lambda>0$,
where ${\rm Tr}(L)$ is the trace of the operator $L$. Our final  assumption describes the property of $K$ via the effective dimension.

\begin{assumption}\label{Assumption:effective dimension}
 There exists an $s\in(0,1]$ such that
\begin{equation}\label{assumption on effect}
      \mathcal N_t(\lambda)\leq C_0\lambda^{-s},\qquad \forall ~ t=1,2,\dots,T,
\end{equation}
where $C_0\geq 1$ is  a constant independent of $\lambda$.
\end{assumption}

From the definition of $\mathcal{N}_t(\lambda)$, \eqref{assumption on effect} always holds for  $C_0=\kappa:=\max_{t=1,\dots,T}\sqrt{\sup_{x\in\mathcal X_t} K(x,x)}$ and $s=1$.
Therefore, without this assumption, our results below always hold for $s=1$. This assumption is introduced to encode the property of the kernel in our theoretical analysis to make the established generalization error bounds more delicate and has been  utilized in studies on kernel learning \citep{lin2017distributed,lin2018distributed}.

In summary, four assumptions concerning $D$,  $P_t$, $Q^*_t$, and $K$ are involved in our analysis. These assumptions can be simultaneously satisfied easily. {For example, assume the state space is $\mathcal S_t:=[0,1]$ and the action space is $\{0,1\}$; conditioned on $s_{1:t},a_{1:t-1}$, one adopts the decision $a_t=1$ with probability $1/2$ and the reward function $R_t(x) \equiv C$ for some constant $C$. If we adopt the Sobolev kernel $K(x_t,x'_t)=\max\{x_t\cdot x_t', 1-x_t\cdot x_t' \}$, then
the above assumptions hold simultaneously with $\mu=2$, $M=C$, $r=1$, and $s=1$. }
Based on the above assumptions, we bound the generalization error of KRR-DTR to show the power of kernel-based Q-learning in the following theorem.

\begin{theorem}\label{Theorem:KRR-generalization}
Under Assumptions 1-4, with $\frac12\leq r\leq 1$ and $0<s\leq 1$, if $\lambda_1=\dots=\lambda_{T}=|D|^{-\frac{1}{2r+s}}$, then
\begin{equation}\label{Generalization-error-KRR}
   E\left[V_1^*(S_1)-V_{ {{\pi}_{D,\vec{\lambda}},1}}(S_1)\right]
  \leq
  C_1(T,\mu)|D|^{-\frac{r}{2r+s}},
\end{equation}
where
$
  C_1(T,\mu):= C_1
(2\mu\bar{C})^TT
     \sum_{t=1}^T 2^{-t}  \sum_{\ell=t}^T      \prod_{k=\ell}^{T-1}\left((T-k+2)(2\mu^{1/2})^{k-\ell}+1\right)
$,
and $\bar{C}$ and $C_1$ are constants depending only on $M$, $C_0$,  $\kappa$, $r$, $s$, and $\max_{t=1,\dots,T}\|h_t\|_{\mathcal L_t^2}$.
 \end{theorem}

Theorem \ref{Theorem:KRR-generalization} quantifies the relation between prediction accuracy of KRR-DTR and  the data size. According to \eqref{Generalization-error-KRR}, the generalization error of KRR-DTR decreases with respect to the data size. Moreover, it depends heavily on the regularity of the optimal Q-functions and the property of the adopted kernel. In particular, the better the regularity of $Q_t^*$, the smaller the generalization error; the smaller the effective dimension of the kernel, the better the quality of KRR-DTR. The derived rate $\mathcal O\left(|D|^{-\frac{r}{2r+s}}\right)$ is the same as that for the classical KRR for regression in supervised learning \citep{caponnetto2007optimal,lin2017distributed} under similar assumptions. The derived generalization error is dimensionality independent, showing that using kernel methods in Q-learning is essentially better than linear approaches.

In Theorem \ref{Theorem:KRR-generalization}, the constant $C_1(T,\mu)$ behaves exponentially with respect to the horizon $T$,  {which is different from the results in \citep{fan2020theoretical}}. The main reason is that, except for the mild restriction in Assumption \ref{Assumption:case-1}, we do not impose any restrictions on the distribution $P$, unlike \citep{fan2020theoretical}. This property makes our analysis suitable only for Q-learning with a small $T$, coinciding  with the practical implementation requirements of DTR in Table \ref{Tab:setting}. Additional restrictions on $P$, such as the concentration-type assumption in \citep{fan2020theoretical} and margin-type assumption in \citep{oh2022generalization}, can improve the constant $C_1(T,\mu)$ and the convergence rate $|D|^{-\frac{r}{2r+s}}$. However, as these assumptions are difficult to satisfy in practice, we do not use them in the present paper.

The derived generalization error bounds in \eqref{Generalization-error-KRR} are much better than those for linear Q-learning  \citep{murphy2005generalization,oh2022generalization} in three aspects. First, we only impose the regularity assumption \eqref{regularitycondition} on optimal Q-functions, whereas  \citep{murphy2005generalization}  and \citep{oh2022generalization} required $Q_t^*$ to belong to a linear combination of finite basis functions.
Second, the derived convergence rate $|D|^{-r/(2r+s)}$ is better than those established for linear Q-learning in previous studies, where rates slower than $|D|^{-1/4}$ were derived under stricter assumptions. For sufficiently small values of $s$, our derived rates can be of the order $|D|^{-1/2}$.
Finally, due to linearity, the rates established in \citep{murphy2005generalization} and \citep{oh2022generalization} depend heavily on the dimension $d$, whereas our results are dimensionality independent, showing the power of kernelization.

Our next theoretical result presents a solid theoretical guarantee for the feasibility of DKRR-DTR.


\begin{theorem}\label{Theorem:DKRR-generalization}
Under Assumptions 1-4, with $\frac12\leq r\leq 1$ and $0<s\leq 1$, if $\lambda_1=\dots=\lambda_{ {T}}=|D|^{-\frac{1}{2r+s}}$, $|D_1|=\dots=|D_m|$, and
\begin{equation}\label{rest.m}
        m (\log m+1)\leq \frac{|D|^{\frac{2r+s-1}{4r+2s}}} {\log|D|},
\end{equation}
then
\begin{equation}\label{Generalization-error-DKRR}
   E\left[V_1^*(S_1)-V_{  { {\overline{\pi}_{D,\vec{\lambda}},1}} }(S_1)\right]
  \leq
   C_2(T,\mu)|D|^{-\frac{r}{2r+s}},
\end{equation}
where
$  C_2(T,\mu):= C_2
(2\mu\hat{C})^TT
     \sum_{t=1}^T 2^{-t}  \sum_{\ell=t}^T      \prod_{k=\ell}^{T-1}\left((T-k+2)(2\mu^{1/2})^{k-\ell}+1\right)
$,
and $\hat{C}$ and $C_2$ are constants depending only on $M$, $C_0$,  $\kappa$, $r$, $s$, and $\max_{t=1,\dots,T}\|h_t\|_{\mathcal L_t^2}$.
\end{theorem}

Theorem \ref{Theorem:DKRR-generalization} presents a sufficient condition of $m$ to guarantee the excellent generalization performance of DKRR-DTR. In particular, if the number of data subsets $m$ and  {the total number of samples $|D|$} satisfy the relation \eqref{rest.m}, then DKRR-DTR performs almost the same as KRR-DTR, showing that the computation reduction approach made by distributed learning does not degrade the generalization performance of KRR-DTR. The restriction on $m$, such as \eqref{rest.m}, is necessary because the generalization performance of KRR-DTR cannot be realized if we use DKRR-DTR with each data subset containing only a few samples.

Theorem \ref{Theorem:DKRR-generalization} shows that DKRR-DTR possesses all the advantages of KRR-DTR,  {but its training cost is only $\mathcal O(|D|^3/m^3)\times m=\mathcal O(|D|^3/m^2)$}, which is $1/m^2$ times smaller than that of KRR-DTR. Given that the quality of Q-learning algorithms is measured by their generalization performance, computational costs, and stability, the above assertions show that DKRR-DTR is better than linear Q-learning in generalization, better than deep Q-learning and KRR-DTR in computation, and better than deep Q-learning in stability as the optimization model defined by \eqref{DKRR-Q-Local-T} can be analytically solved, whereas obtaining global minima with high-quality deep neural networks remains an open question in the realm of deep learning \cite[Chap.8]{goodfellow2016deep}. These findings illustrate that DKRR-DTR is a promising approach to yielding high-quality DTRs.

\section{Simulations}\label{sec.simulation}

  {In this section, two types of clinical trials for cancer treatment are simulated to demonstrate the efficacy of the proposed method compared with fitted-Q-iteration methods, where the hypothesis spaces for Q-function approximation are linear spaces and deep neural networks, denoted by LS-DTR and DNN-DTR, respectively, as well as several commonly used policy gradient-related methods, which include policy gradient (PG) \citep{Sutton1999PG}, actor-critic (AC) \citep{Mnih2016AC}, soft actor-critic (SAC) \citep{Haarnoja2019SAC}, proximal policy optimization (PPO) \citep{Schulman2017PPO}, deep deterministic policy gradient (DDPG) \citep{Lillicrap2016DDPG}, and twin-delayed deep deterministic (TD3) policy gradient \citep{Fujimoto2018TD3}. Additionally, we compare the proposed method with several offline methods, specifically model-based policy optimization (MBPO) \citep{Michael2019}, using base agents PPO, DDPG and TD3, respectively.
For a fair comparison, we clarify some implementation details in the following:
\begin{itemize}
\item In fitted-Q-iteration, consistent hyperparameter values are maintained across clinical stages. Kernel-based Q-learning employs the Gaussian kernel. DKRR-DTR distributes training samples randomly and equally across local machines. DNN-DTR utilizes a fully connected feedforward network with sigmoid activation and Xavier initialization \citep{Glorot2010}.
\item For PG-related methods, a fully connected feedforward neural network with the ReLU activation function and Kaiming initialization \citep{He2015Kaming} is employed to approximate both the policy and value functions.
\item For MBPO methods, we first train an environment model using offline experiences. This involves training DNNs to approximate transition, reward, and terminal functions. Here, we use three transition functions as an ensemble of transition models to generate experiences. Subsequently, we update the actor and critic using the experiences generated by the trained environment model.
\end{itemize}
}

All simulations are run on a desktop workstation equipped with an Intel(R) Core(TM) i9-10980XE 3.00 GHz CPU, an Nvidia GeForce RTX 3090 GPU, 128 GB of RAM, and Windows 10; the DNN-DTR programs are executed on the GPU, and the others are executed on the CPU.
The results are recorded by averaging the results from multiple individual trials with the best parameters.

\subsection{Clinical trial with a small number of treatment options} \label{Clinical_Sim}

  {NSCLC patients typically receive 1-3 treatment lines, flexible depending on disease progression and therapy tolerance.  This simulation aims to develop an optimal treatment policy that maximizes the expected survival time. We use the cancer dynamics model by \cite{Goldberg2012} with aggressive ($A$) and conservative ($B$) treatments to generate the clinical trajectory. Trajectory generation details are given in the appendix.
The following cases are considered to examine the performance of the proposed method:
\begin{itemize}
  \item Single feature $+$ Separately (S+S): The state is the wellness at the start of the current stage, and the Q-functions are estimated separately for actions $A$ and $B$.
  \item Multiple feature $+$ Separately (M+S): The state is comprised of the wellness at the start of the current stage and the reward obtained from the previous stage. The Q-functions are estimated separately for actions $A$ and $B$.
  \item Multiple feature $+$ Jointly (M+J): The state is the same as in the case of M+S, and the Q-function is estimated to be a joint function of the state and action.
  \item Non-Markovian $+$ Jointly (N+J): The state is the same as in the case of M+S, and the Q-function is estimated to be a joint function of all state-action pairs across all stages up to the current one.
\end{itemize}
}

We generate a dataset of $N=10000$ trajectories for training and test the policy $\pi=({\pi}_1,{\pi}_2,{\pi}_3)$ by generating $1000$ new trajectories in which the treatment of the $i$-th stage accords with $\pi_i$. Eight fixed treatments, where $\pi=(a_1,a_2,a_3)$ for $a_i\in\{A,B\}$, are used for baseline. For KRR-DRT and DKRR-DRT, the regularization parameter $\lambda$ is chosen from the set
$\left\{\frac{1}{2^q}|\frac{1}{2^q}>\frac{1}{2N}, q=0,1,2,\cdots\right\}$,
and the kernel width $\sigma$ is chosen from a set of 20 values drawn in an equally spaced logarithmic interval $[0.001,1]$.
 {For DNN-DRT and PG-related methods, the number of hidden layers and the neuron number in each hidden layer are chosen from the sets $\{1,2,3,4\}$ and $\{10,20,\cdots,100\}$, respectively. Additionally, the number of training epochs for DNN-DRT and the maximum number of episodes for PG-related methods are both chosen from the set $\{2000,4000,\cdots,10000\}$.}
The expected value of survival time is estimated using the mean of the survival time of the $1000$ patients. For each method with the best parameters, we conduct the simulation $500$ times and record the average value of the estimated mean survival time for evaluation.\protect\footnotemark[1]
\footnotetext[1]{  {The DNN-DTR and PG-related methods are} conducted $100$ times because of the high training cost.}

\begin{figure*}[t]
    \centering
    \subfigcapskip=-10pt
    \subfigure[Comprehensive comparison of survival time]{\includegraphics[width=16.5cm,height=4.1cm]{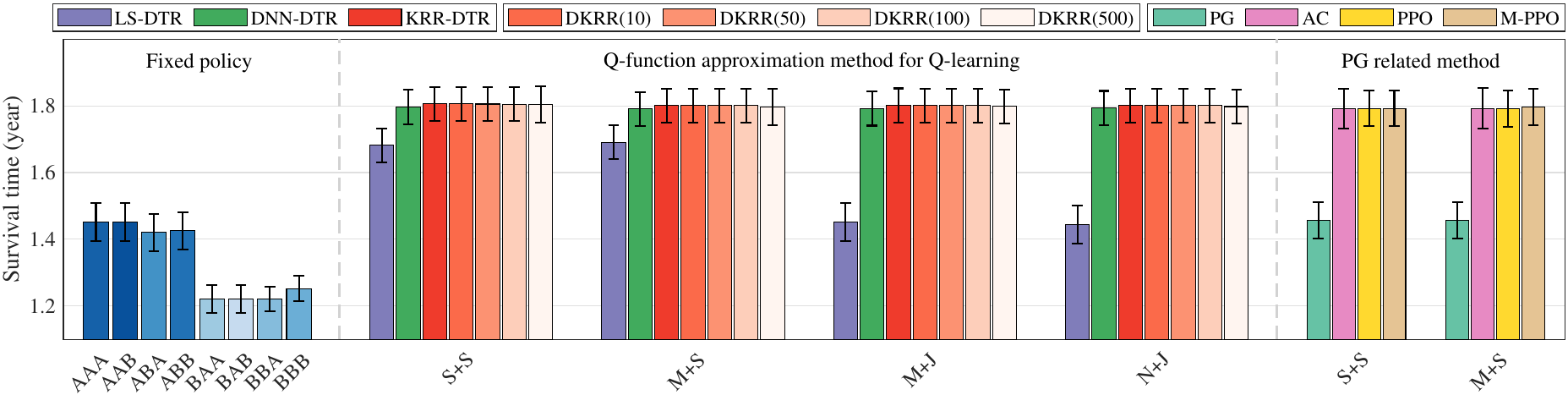}}\\
    \vspace{-0.05in}
    \subfigure[Detailed comparison of survival time]{\includegraphics[width=5.8cm,height=3.7cm]{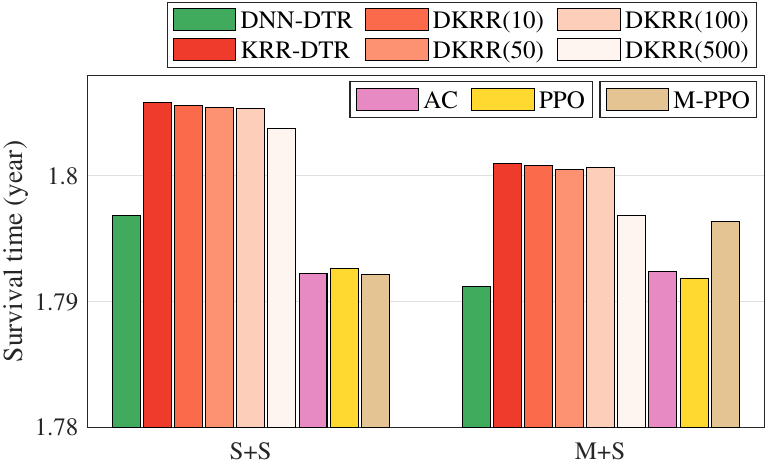}}\hspace{-0.05in}
    \subfigure[Comprehensive comparison of training time]{\includegraphics[width=10.7cm,height=3.65cm]{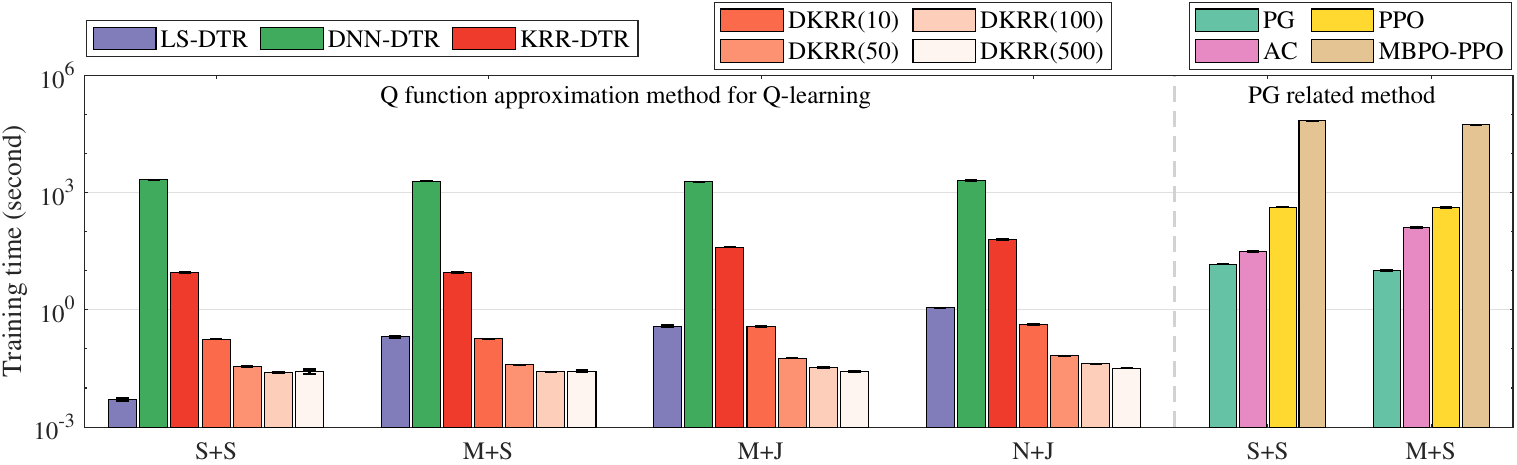}}
	\caption{Comparisons of Survival time and training time for the mentioned methods}\label{Clinical_baseline}
\end{figure*}

  {The comprehensive comparisons of survival time and training time are illustrated in Figure \ref{Clinical_baseline}, where Figure \ref{Clinical_baseline}(b) offers a more detailed examination of the survival times among several policies that exhibit closely comparable performance. DKRR-DTR is abbreviated as DKRR for brevity and the numbers in brackets after the notation DKRR are the numbers of local machines. MBPO using base agents PPO is abbreviated as M-PPO. It should be noted that, when dealing with discrete action spaces, the PPO algorithm calculates the Q-functions separately for each individual action. Furthermore, the identity of the Q-function across all stages necessitates that the state dimensions remain the same at each stage, thus eliminating the non-Markovian case. Consequently, the PPO algorithm involves only two cases regarding the state space: the previously defined single feature and multiple feature, denoted as PPO(S) and PPO(M), respectively. From the above results, we can conclude the following:
1) The policies AAA and AAB have the best survival time among the eight fixed treatments, but they are still significantly worse than the policies obtained by RL methods, which illustrates the power of RL in discovering effective regimens. 2) Apart from DKRR-DTR, LS-DTR has the shortest training time, which increases from case S+S to case N+J. This is because the training complexity of LS-DTR is
proportional to the cube of the data dimension
(the dimension increases from case S+S to case N+J). Nevertheless, LS-DTR has the worst survival time among all RL methods. The survival times of LS-DTR for cases S+S and M+S are obviously higher than those of LS-DTR for cases M+J and N+J, which indicates that LS is not good at jointly handling states and actions when the size of the action space is small. 3) KRR-DTR exhibits the best performance in survival time. Although KRR-DTR does not demonstrate a notable advantage in training time compared to PG, AC, and PPO, it is crucial to highlight that the training procedures of PG, AC, and PPO necessitate direct interaction with the real environment, rendering them impractical for medical diagnosis applications where safety requirements are exceptionally rigorous. This is due to the potential for any exploratory treatment to cause irreparable harm to patients. 4) M-PPO attains comparable survival time outcomes to KRR-DTR by developing an environmental model that circumvents direct interaction with the real environment. However, training the environmental model is extremely time-consuming, requiring training times that are thousands of times longer than that of KRR-DTR. Regarding DNN-DTR, which also eliminates direct interaction with the real environment, its time cost is significantly higher-ranging from tens to hundreds of times greater-than that of KRR-DTR, due to the necessity of training a deep neural network at each stage. 5) DKRR-DTR maintains almost equivalent performance to KRR-DTR regarding survival time. Even with 500 local machines, the performance of DKRR-DTR remains superior to that of DNN-DTR and PG-related methods.  However, in terms of training time, the framework based on distributed computing significantly reduces the training time of KRR-DTR. When the number of local machines is larger than 10, the training time of DKRR-DTR (in all cases except S+S) is even less than that of LS-DTR, whose training complexity is independent of training data size.  Overall, DKRR-DTR is highly efficient in the training stage while generalizing comparably to state-of-the-art methods, which implies that DKRR-DTR can be applied to large-scale DTR problems.}

\subsection{Clinical trial with a large number of treatment options}\label{TrialContinuous}

  {This simulation considers a fixed-length clinical trial with monthly treatment stages. Each patient receives a treatment at the start of each month. We design an optimal strategy balancing drug efficacy and toxicity to maximize cumulative survival probability (CSP) after all courses. The model proposed by \cite{Zhao2009} is used to generate clinical data. Details are in the appendix. Note that the action space is continuous in this simulation. The aforementioned PG-related methods are capable of directly handling such continuous action spaces. However, for fitted Q-iteration methods such as LS-DTR, DNN-DTR, and KRR-DTR, we must discretize the continuous action space into numerous dose levels. Consequently, the following cases are the sole focus of our consideration:
\begin{itemize}
\item Markovian $+$ Jointly (M+J): The state consists of the toxicity and tumor size at the current stage, and the Q-function is estimated to be a joint function of the state and action.
\item Non-Markovian $+$ Jointly (N+J): The state is the same as in the case of M+J, and the Q-function is estimated to be a joint function of all state-action pairs across all stages up to the current stage.
\end{itemize}
\noindent Note that PG-related methods calculate the Q-functions as a combined function of both the state and action, and the identity of the Q-function across all stages negates the non-Markovian case.}

The settings of this simulation are as follows. We generate $N=20000$ trajectories for training and  $1000$ trajectories for evaluation. Ten fixed treatments whose dose levels range from $0.1$ to $1.0$ in increments of $0.1$ are used for baselines. For KRR-DRT and DKRR-DRT, $\lambda$ is chosen from the set $\{\frac{1}{2^q}|\frac{100}{N}>\frac{1}{2^q}>\frac{1}{10N}, q=0,1,2,\cdots\}$,
and the kernel width $\sigma$ is chosen from a set of 20 values drawn in an equally spaced logarithmic interval $[0.01,10]$.  {The parameter ranges for DNN-DRT and PG-related methods follow Section \ref{Clinical_Sim}}. The expected CSP value is estimated using the mean CSP value of the $1000$ patients. The simulation is conducted $100$ times for each method with the best parameters, and the average value of the estimated mean CSP is reported.\protect\footnotemark[2]
\footnotetext[2]{The DNN-DTR and PG-related methods are conducted $20$ times.}


\begin{figure*}[t]
    \centering
    \subfigcapskip=-10pt
    \subfigure[Comprehensive comparison of CSP]{\includegraphics[width=16cm,height=4.2cm]{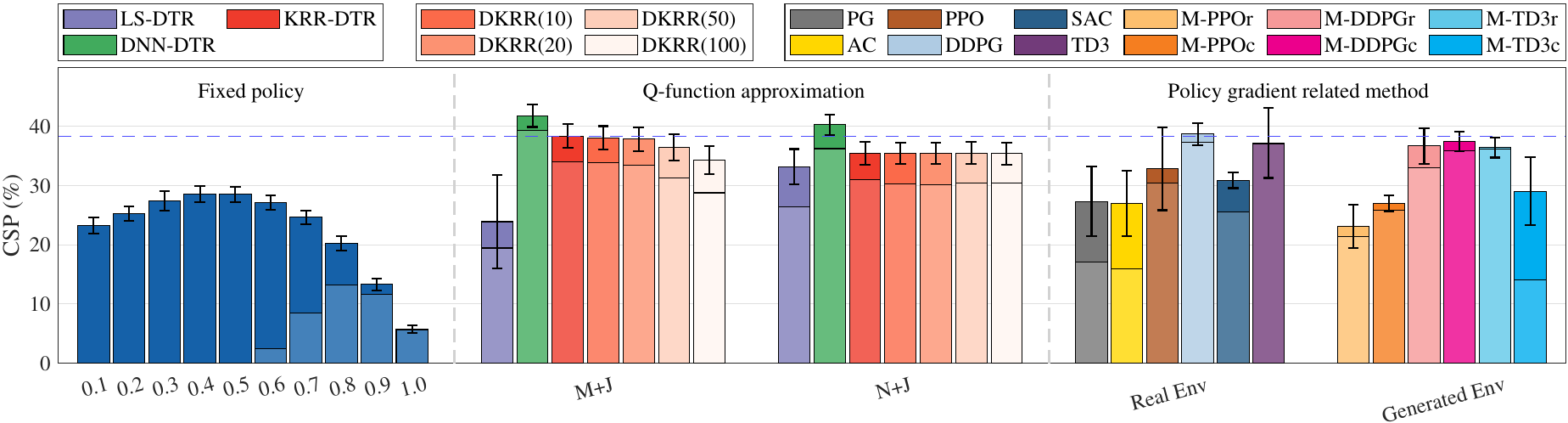}}\\
    \vspace{-0.08in}
    \subfigure[Comprehensive comparison of training time]{\includegraphics[width=12.5cm,height=4.2cm]{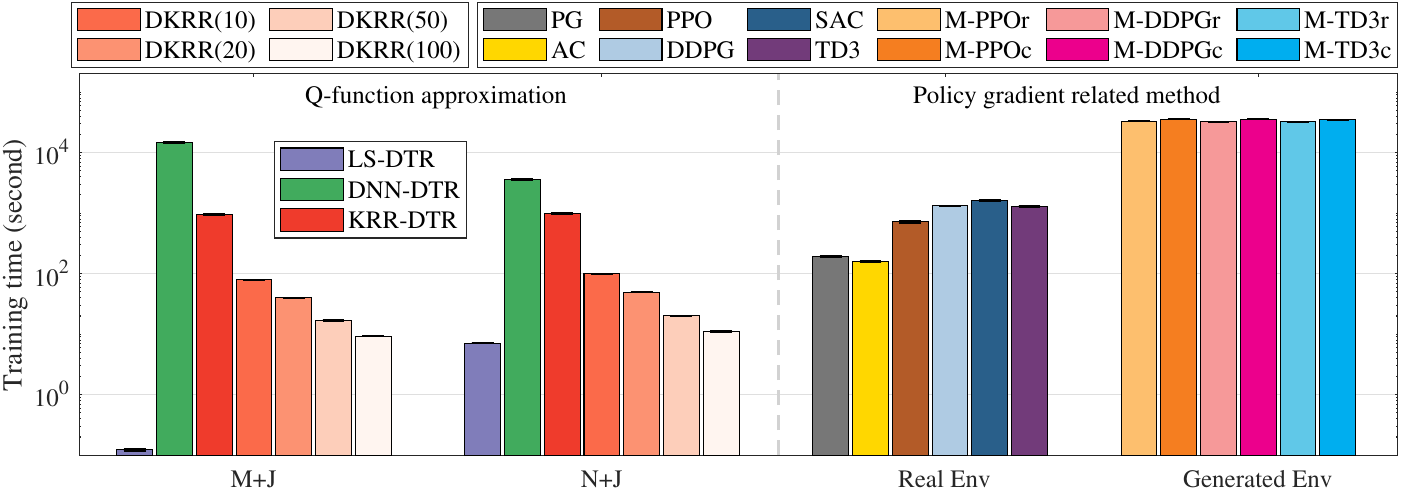}}
	\caption{Comparisons of CSP and training time for the mentioned methods.\label{Cancer_baseline}}
\end{figure*}

Comparisons of the CSP and training time are shown in Figure \ref{Cancer_baseline}, where the light and dark shades of each color series represent the cumulative cured probability (CCP) and trial end probability (TEP) of the compared methods, respectively.   {The dashed blue line depicted in Figure 4(a) signifies the CSP of KRR-DTR in the case of M+J. In MBPO methods, the suffix `c' and `r' denote the utilization of classification and regression techniques, respectively, for the construction of the reward model.} Here, the CCP is the proportion of patients cured in the treatment cycle, and TEP is the proportion of patients alive but not cured at the end of the trial.
From the results, we obtain the following observations: 1) Some fixed treatments, such as dose levels of $0.3-0.6$, have good CSP values and even outperform LS-DTR in the case of M+J. However, their CCP values are evidently lower than those of Q-learning methods. The CCP is usually more important than TEP because patients prefer to be cured in the treatment cycle.
Hence, RL can provide effective treatment policies for life-threatening diseases. 2)   {DNN-DTR performs the best because of its excellent nonlinear fitting capability for Q-functions, but its training is very time-consuming.} LS-DTR has the shortest training time; the CSP value of LS-DTR in the case of N+J is significantly better than that in M+J due to the expansion of the linear hypothesis space (the dimension of data input grows with the number of stages in N+J), but LS-DTR still has the lowest
CSP values and the highest standard deviations.   {3) The training time of PG-related methods that interact with the real environment primarily depends on the number of episodes, and it is either comparable to or superior to that of KRR-DTR. Despite that DDPG, a method within this category, achieves superior performance on CSP compared to KRR-DTR, such methods are
unsuitable for application in medical diagnosis due to their inability to meet the extremely high safety requirements.
MBPO methods do not require interaction with the real environment during training and demonstrate nearly equivalent performance to KRR-DTR in terms of CSP. However, the necessity to train an environmental model significantly prolongs their overall training time compared to all other methods.  4) KRR-DTR is a compromise between the CSP and training time; its CSP value is slightly smaller than that of DNN-DTR, but it is more efficient than DNN-DTR and MBPO methods. More importantly, it possesses the capability of offline training, a feature that fully meets the stringent requirements for safety. By distributing data across multiple local machines, DKRR-DTR effectively circumvents the limitations of KRR-DTR in handling large-scale data training. Its parallel computing capability significantly reduces the training time, while maintaining comparable performance levels to KRR-DTR in terms of CSP.}

These simulations illustrate that DKRR-DTR has higher efficiency and superior performance in dealing with large-scale DTR problems compared with state-of-the-art methods.

\section{Conclusion}

  {This paper introduced a distributed Q-learning algorithm that integrates DKRR with Q-learning to mitigate the computational burden associated with DTRs. Through the development of a novel integral operator, we established tight generalization error bounds for the proposed algorithm within the framework of statistical learning theory. Our theoretical findings indicate that the distributed operator does not compromise the generalization error, provided that the data are not excessively fragmented into numerous subsets. Furthermore, the derived error bounds are independent of the data dimension, implying that the proposed algorithm scales effectively with respect to the problem's dimensions.
In addition to our theoretical analysis, we also applied the proposed algorithm to two types of clinical trials for cancer treatment, including those aimed at increasing survival time and CSP. The numerical results verify our theoretical assertions and demonstrate the feasibility and efficiency of the proposed algorithm in applications.}

\ACKNOWLEDGMENT{  We are   deeply grateful to Professor Shaojie Tang for his insightful suggestions and significant contributions to this  work. The work of this paper is partially supported  by National Natural Science Foundation of China (Grant Nos. 12471486,12371513,2276209). }

\newpage

\begin{APPENDICES}

\section{Introduction of Simulation 1}\label{IntroSim1}
This section provides a thorough description of the trajectory data production for Section 5.1.
The observation period for patients in the clinical trial is 5 years. At each time point $t\in[0,5]$, the tumor size denoted by $M(t)$ and the wellness denoted by $W(t)$ for each patient are recorded.
When a patient's wellness is less than 0.2, the patient suffers from death. When a patient's tumor size reaches 1,
treatment should be administered. Here, an aggressive treatment (denoted by $A$) or a conservative treatment (denoted by $B$) is considered for application to patients. The immediate effects on wellness and tumor size for the two treatments are formulated as
\begin{align*}
 & W(t_i^+|A) = W(t_i) - 0.5, \quad M(t_i^+|A) = 0.1 M(t_i)/W(t_i), \\
 & W(t_i^+|B) = W(t_i) - 0.25, \quad M(t_i^+|B) = 0.2 M(t_i)/W(t_i).
\end{align*}
It can be seen that the aggressive treatment $A$ yields a greater decrease in tumor size and wellness than the conservative treatment $B$. In the following, we rewrite both of $W(t_i^+|A)$ and $W(t_i^+|B)$ as $W(t_i^+)$, and rewrite both of $M(t_i^+|A)$ and $M(t_i^+|B)$ as $M(t_i^+)$ for convenience. With the treatment, a patient's current survival time $\tau_i$ is drawn independently according to an exponential distribution, with a mean of $0.15(W(t_i^+)+2)/M(t_i^+)$. The cancer dynamics after $t_i$ satisfy the following equations:
\begin{equation*}
W(t) = W(t_i^+) + (1-W(t_i^+))(1-2^{-(t-t_i)/2}), \qquad  M(t) = M(t_i^+) + 2M(t_i^+)(t-t_i).
\end{equation*}
Noted that the $i$th stage begins at time point $t_i$, and ends at time point $t_{i+1}$ such that $M(t_{i+1})=1$ for some $t_i<t_{i+1}<5$, or the patient dies, or the clinical trial ends. Specifically, we can obtain the critical time point
$\hat{t}_i = t_i + 0.75\left(\frac{1-M(t_i^+)}{M(t_i^+)} \right)$
from $M(\hat{t}_i)=1$, and thus compute the end of the $i$th stage via $t_{i+1}=\min\{\hat{t}_i, t_i+\tau_i, 5\}$.

We assume that only patients in urgent need of treatment are enrolled in the clinical trial, which means that all patients meet $M(0)=1$ and thus $t_1=0$. The wellness of patients at the beginning of the first stage is drawn independently according to the uniform distribution in the interval $[0.5,1]$. A treatment is randomly chosen from the set $\{A,B\}$ for each time point $t_i$. The reward is defined as the actual survival time of the current stage. Due to the differences in initial wellness, survival time, and treatment policies for different patients, the generated trajectories may have different numbers of stages. Here, the total number of stages for all patients is limited to $3$. For the missing stages of the trajectory data with a number of stages less than 3, we fill wellness and reward as zeros and randomly choose actions from the set $\{A,B\}$ with equal probability, modifying the trajectory to a fixed length of $3$.

\section{Introduction of Simulation 2}\label{IntroSim2}
This section describes the process of generating trajectory data in Section 5.2.
In this simulation, patients are monitored monthly for 6 months, and treatment is applied to each patient at the beginning of each month. The relation of a drug's toxicity $W_i$, tumor size $M_i$, and drug dosage $A_i$ of treatment is characterized by the following system of ordinary difference equations:
\vspace{-0.2in}
\begin{align*}
& W_{i+1}=W_i+0.1(M_i\vee M_0)+1.2(A_i-0.5), \\
& M_{i+1}=\left(M_i+\left(0.15(W_i\vee W_0)-1.2(A_i-0.5)\right) \times \bm{1}_{M_i>0}\right)\vee 0,
\end{align*}
for $i=1,\cdots,6$, where
$\bm{1}_{M_i>0}$ is an indicator function representing that there will be no future recurrence of the tumor when the tumor size reaches 0 (that is, the patient has been cured). The initial values of toxicity $W_1$ and tumor size $M_1$ are drawn independently according to the uniform distribution on the interval $(0,2)$.
The drug dosage treatment sets for the first stage and last five stages are $\{0.51,0.52,\cdots,1\}$ and $\{0.01,0.02,\cdots,1\}$, respectively. They are discretized from drug dosage intervals $(0.5,1]$ and $(0,1]$ with an increment of size 0.01. The action $A_1$ and the actions $A_i$ ($i=2,\cdots,6$) are randomly chosen from the sets $\{0.51,0.52,\cdots,1\}$ and $\{0.01,0.02,\cdots,1\}$ with equal probability, respectively.
To increase the stochasticity of the survival status, we define the conditional probability of death for the $i$th stage as
\begin{equation*}\label{pro_survival}
p_i = 1-\exp\left( -\exp(W_i+M_i-4.5)\right).
\end{equation*}
The survival status (death is coded as 1) is drawn independently according to the Bernoulli distribution $B(p_i)$. The reward for each stage is assumed to depend on the states (including toxicity and tumor size) observed right before and after each action, and it can be decomposed into three types of rewards: $R_{i,1}(A_i,W_{i+1},M_{i+1})$ related to survival status, $R_{i,2}(W_i,A_i,W_{i+1})$ related to toxicity change, and $R_{i,3}(M_i,A_i,M_{i+1})$ related to tumor size change. Specifically, they are defined by
\begin{equation}\label{Rt1}
R_{i,1}(A_i,W_{i+1},M_{i+1}) = -6,  \quad \mbox{if patient died},
\end{equation}
otherwise,
\begin{align}
& R_{i,2}(W_i,A_i,W_{i+1})=\left\{
\begin{array}{lcl}
0.5,       &\quad \mbox{if}     & W_{i+1}-W_i \leq -0.5,\\
-0.5,      &\quad \mbox{if}     & W_{i+1}-W_i \geq 0.5,\\
0,         &\quad \mbox{if}     & -0.5 < W_{i+1}-W_i < 0.5,
\end{array} \right.\label{Rt2}\\
& R_{i,3}(M_i,A_i,M_{i+1})=\left\{
\begin{array}{lcl}
1.5,       &\quad \mbox{if}     & M_{i+1}=0,\\
0.5,       &\quad \mbox{if}     & M_{i+1}-M_i \leq -0.5 ~ \mbox{and} ~ M_{i+1}> 0, \\
-0.5,      &\quad \mbox{if}     & M_{i+1}-M_i \geq 0.5,\\
0,         &\quad \mbox{if}     & -0.5 < M_{i+1}-M_i < 0.5.
\end{array} \right.\label{Rt3}
\end{align}
Because overall survival is the main focus of clinical interest, we take a high penalty of $-6$ for the death of a patient in the equation (\ref{Rt1}), and we take a relatively large bonus of $1.5$ for the cure of a patient in the equation (\ref{Rt3}). In other cases, rewards are given according to the changes in toxicity and tumor size in two consecutive stages.

\section{An Analysis of the Impact of Stage Number on DTR Performance}
In this section, we present a comprehensive analysis of the impact of stage number $T$ on DTR performance. To this end,
we first elucidate several implementation details as follows:
\begin{itemize}
\item For Simulation 1, based on previous results, we select the M-PPO algorithm, which performs best among the policy gradient-related algorithms, and the best fixed policy to compare with our proposed KRR-DTR. The number of stages $T$ varies from the set $\{3,4,\cdots,10\}$. For each given $T$, the observation period for patients in the clinical trial is $T+2$ years.
The immediate effects on wellness for the two treatments are formulated as
\begin{equation*}
W(t_i^+|A) = W(t_i) - (0.53-0.01T), \quad W(t_i^+|B) = W(t_i) - (0.265-0.005T).
\end{equation*}
With the treatment, a patient's current survival time $\tau_i$ is drawn independently according to an exponential distribution, with a mean of $$
(0.12+0.01*T)(W(t_i^+)+2)/M(t_i^+).
$$
The tumor size after $t_i$ satisfies the following equations:
\begin{equation*}
 M(t) = M(t_i^+) + (1.7+0.1T)M(t_i^+)(t-t_i).
\end{equation*}
The remaining parameter settings and environmental dynamics are the same as those in the manuscript.

\begin{figure}[t]
	\centering
	\subfigure{
    	\includegraphics[width=6cm,height=5cm]{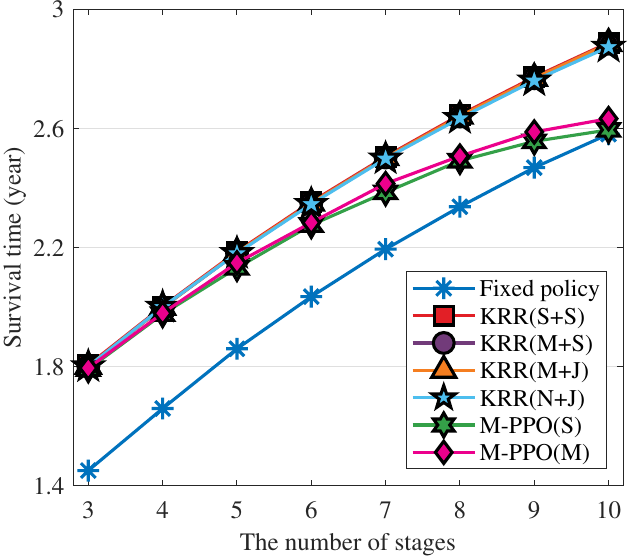}
	}\hspace{-0.05in}
		\subfigure{
		\includegraphics[width=6cm,height=5cm]{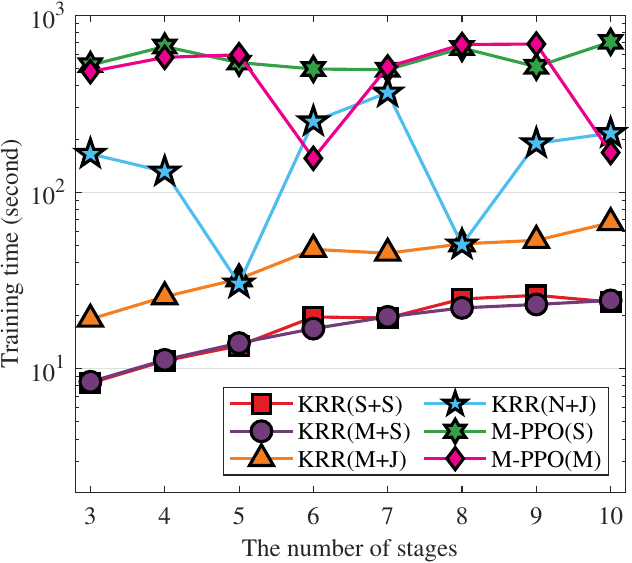}
	}
	\caption{Comparison of survival time and training time with increasing number of stages for the methods of fixed policy, KRR-DTR, and MBPO-PPO}
	\label{NtrVaryHSim1}
\end{figure}
\begin{figure}[t]
	\centering
	\subfigure{
    	\includegraphics[width=6cm,height=4.8cm]{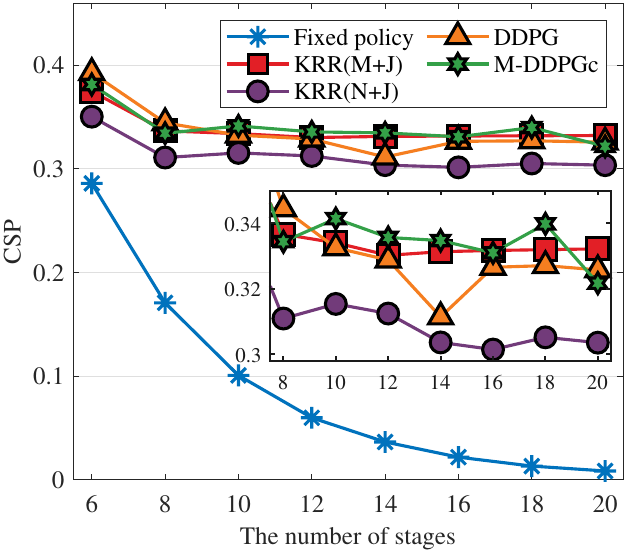}
	}\hspace{-0.05in}
		\subfigure{
		\includegraphics[width=6cm,height=4.8cm]{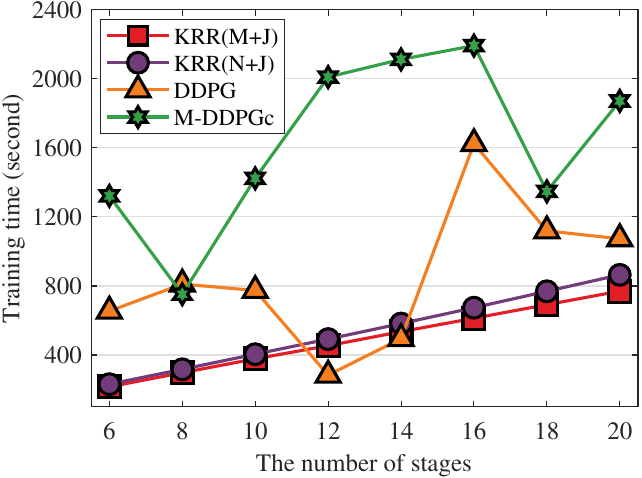}
	}
	\caption{Comparison of CSP and training time with increasing number of stages for the methods of fixed policy, KRR-DTR, DDPG and MBPO-DDPG}
	\label{NtrVaryHSim2}
\end{figure}

\item For Simulation 2, we choose the two most prominent methods within policy gradient approaches: DDPG and M-DDPGc for comparison with our proposed method. Furthermore, we select the optimal constant dose regimen as a benchmark for reference purposes. The number of stages $T$ varies within the set $\{6,8,\cdots,20\}$. For each specific stage number $T$, the parameter configurations and environmental dynamics remain consistent with those detailed in the manuscript.

\item For policy gradient-related methods, in the context of both the policy network and the value network, the number of hidden layers, the neuron number in the hidden layers, and the critical threshold for terminating the training are selected from the sets $\{1, 2, 3, 4\}$, $\{10, 20, \cdots, 100\}$, and $\{2000, 4000, \cdots, 10000\}$, respectively.

\item For each specified stage number $T$, we generate 10000 trajectories for training in our proposed KRR-DTR. To ensure fairness, in the context of model-based policy optimization (MBPO) related methods, we utilize 10000 episodes of data obtained through real environmental interactions to train the environment model, which includes the transition function, reward function, and terminal function. Furthermore, the maximum number of training episodes for the agents employed in policy gradient-related methods is set to 10000. All compared methods are evaluated on newly generated set of 1000 trajectories.  The results are recorded by calculating the average of multiple individual trials with the best parameters.
\end{itemize}


The comparisons of the specific targets of different tasks (survival time in Simulation 1 and CSP in Simulation 2) and the training time, with an increasing number of stages for the involved methods, are shown in Figures \ref{NtrVaryHSim1}-\ref{NtrVaryHSim2}, where KRR-DTR is abbreviated as KRR for brevity and the symbols in brackets represent the different considered cases. It should be noted that, when addressing discrete action spaces, the PPO algorithm calculates the Q-functions separately for each individual action. Moreover, the identity of the Q-function across all stages necessitates that the state dimensions remain the same at each stage, thus eliminating the non-Markovian case. Consequently, the PPO algorithm involves only two cases regarding the state space: one case where the wellness $w_i$ serves as the sole component of the state $s_i$, and another where both the wellness $w_i$ and the reward $r_{i-1}$ obtained from the previous stage collectively constitute the state $s_i$. These two cases are denoted as PPO(S) and PPO(M), respectively. The DDPG algorithm calculates the Q-functions as a combined function of both the state $s_i$ and action $a_i$, and the identity of the Q-function across all stages also negates the non-Markovian case. From the results, we have the following observations:
\begin{itemize}
\item In the task of Simulation 1, which focuses on maximizing the expected survival time, the average survival time of the compared methods consistently increases as the number of treatment stages $T$ rises (equivalently, as the observation period $T+2$ is extended). Although the PPO algorithm performs comparably to our proposed KRR-DTR method when the number of stages is relatively low, as the number of stages progressively increases, the superiority of PPO over fixed policy diminishes. In stark contrast, KRR-DTR demonstrates significant and consistent advantages across all stages.
\item In the task of Simulation 2, which aims at maximizing the CSP after $T$ treatment courses, the average CSP of all compared methods demonstrates a consistent decline as the number of treatment stages increases, attributed to the progressive reduction in the number of survivors. Specifically, the adoption of the fixed policy results in a particularly steep decline in CSP, forming a stark contrast with the more gradual decrease exhibited by strategies obtained through reinforcement learning. As the number of stages reaches 14 or beyond, the CSP under reinforcement learning strategies remains relatively stable, clearly showcasing the superior efficacy of reinforcement learning.
In the `M+J' case, KRR-DTR demonstrates performance comparable to DDPG. However, in the `N+J' case, KRR-DTR is slightly inferior to DDPG, aligning with the related findings presented in the main manuscript. This may stem from the fact that the simulation experiment is grounded on the Markov assumption, rendering the use of non-Markov states inappropriate for training agents. It is noteworthy that DDPG also did not utilize non-Markov states during its training process.
\item In terms of training time, these two types of tasks exhibit certain similarities. Specifically, the KRR-DTR method requires training a corresponding Q-function for each stage, thus its training time roughly increases linearly with the number of stages. Conversely, policy gradient-related methods utilize a consistent Q-function across all stages, with training time primarily contingent upon the quantity of episodes utilized in the training process. The fluctuation in training time primarily arises from the best number of episodes selected from the set $\{2000, 4000,..., 10000\}$,  thereby preventing a proportional increase in training time with the augmentation of stages, as observed in the case of KRR-DTR. It is worth noting that, to more clearly reveal the relationship between training time and the number of stages, the time for training the environment model (both tasks take tens of thousands of seconds) is not included in the training time when considering MBPO-related methods (such as MBPO-PPO and MBPO-DDPGc). Despite this, the training time of KRR-DTR remains significantly shorter compared to that of policy gradient-based algorithms. However, given the linear relationship between the training time of KRR-DTR and the number of stages, KRR-DTR is not suitable for handling DTR tasks that involve hundreds or more stages. While DKRR-DTR can improve the training speed of KRR-DTR to a certain extent through distributed computation, its training time still exhibits a linear relationship with the number of stages. Additionally, an excessively high number of distributed nodes may adversely affect the generalization performance of DKRR-DTR.
\end{itemize}

In summary, our proposed KRR-DTR method is ideally suited for DTR tasks with a small number of stages, specifically in the tens, as shown in the above simulations. However, neither its training efficiency nor generalization performance renders it viable for DTR tasks that encompass hundreds or more stages.

\section{Proofs}
In this section, we develop a novel integral operator approach for Q-learning. Our main tools are error decomposition for KRR-DTR and DKRR-DTR based on their operator representations and uniform bounds for KRR-DTR and DKRR-DTR estimators.


\subsection{Operator representations and differences}
The core of the integral operator approach developed in \citep{smale2005shannon,caponnetto2007optimal,lin2017distributed} is the operator representation of kernel-based learning algorithms, error decomposition, and tight bounds of operator differences.

For each $t=1,\dots,T$, let $S_{D,t}:\mathcal H_{K,t}\rightarrow\mathbb R^{|D|}$ be the sampling
operator defined by
$
         S_{D,t}Q_t:=(Q_t(x_{i,t}))_{i=1}^{|D|}.
$
Its scaled adjoint $S_{D,t}^T:\mathbb R^{|D|}\rightarrow
\mathcal H_{K,t}$  is
given by
$
       S_{D,t}^T{\bf c}:=\frac1{|D|}\sum_{i=1}^{|D|}c_iK_{x_{i,t}}
$
for $K_x=K(x,\cdot)$ and ${\bf
       c}:=(c_1,c_2,\dots,c_{|D|})^T
       \in\mathbb
       R^{|D|}.$
Define $ L_{K,D,t}Q_t:=S_{D,t}^TS_{D,t}Q_t=\frac1{|D|}\sum_{i=1}^{|D|}Q_t(x_{i,t})K_{x_{i,t}}
$.
Then,  $Q_{D,\lambda_t,t}$, defined by (9),
can be analytically represented as \citep{smale2005shannon}
\begin{equation}\label{operator-KRR}
    Q_{D,\lambda_t,t}=\left(L_{K,D,t}+\lambda_t
    I\right)^{-1}S_{D,t}^T\mathbf{y}_{D,t},
\end{equation}
where {$\mathbf{y}_{D,t}:=(y_{1,t},\dots,y_{|D|,t})^T$.}
Define
\begin{equation}\label{def.without-stage}
      \hat{Q}_{D,\lambda_t,t}:=\left(L_{K,D,t}+\lambda_t
    I\right)^{-1}S_{D,t}^T\mathbf{y}^*_{D,t},
\qquad\  \
      Q^\diamond_{D,\lambda_t,t}:=\left(L_{K,D,t}+\lambda_t
    I\right)^{-1}L_{K,D,t}Q^*_t,
\end{equation}
where $\mathbf{y}^*_{D,t}=(y^*_{1,t},\dots,y^*_{|D|,t})^T$.
Then $\hat{Q}_{D,\lambda_t,t}$ can be regarded as running KRR on the data set $\{(x_{i,t},y^*_{i,t})\}_{i=1}^{|D|}$, while $Q^\diamond_{D,\lambda_t,t}$ is a noise-free version of $\hat{Q}_{D,\lambda_t,t}$.
Let
\begin{equation}\label{def.without-stage-DKRR}
      \hat{Q}_{D_j,\lambda_{t},t}:=\left(L_{K,D_j,t}+\lambda_{t}
    I\right)^{-1}S_{D_j,t}^T\mathbf{y}^*_{D_j,t},
\qquad\
      Q^\diamond_{D_j,\lambda_{t},t}:=\left(L_{K,D_j,t}+\lambda_{t}
    I\right)^{-1}L_{K,D_j,t}Q^*_t
\end{equation}
be the corresponding local estimators for $\hat{Q}_{D,\lambda_t,t}$ and $ Q^\diamond_{D,\lambda_t,t}$, respectively. Define further
\begin{equation}\label{Def.Qoplus-new}
   Q^\oplus_{D,\lambda_t,t}= (L_{K, D,t}+\lambda_{t}I)^{-1}S_{D,t}^T \overline{\bf y}_{t}
\end{equation}
as the  batch version of ${Q}_{D_j,\lambda_t,t}$, where $\overline{\bf y}_{t}=(\overline{ y}_{1,t},\dots,\overline{ y}_{|D|,t})^T$ with
\begin{equation}\label{def.ybar}
     \overline{y}_{i,t}:=r_{i,t}+\max_{a_{t+1}\in\mathcal{A}_t}\overline{Q}_{D,{\lambda}_{t+1},t+1}(s_{i,1:t+1}, a_{i,1:t},a_{t+1}).
\end{equation}
It is obvious that $ Q^\oplus_{D,\lambda_t,t}$ is different from $ Q_{D,\lambda_t,t}$ due to their different outputs of data.
Denote
\begin{small}
\begin{equation}\label{def.Q-1111-condtional-exp-U}
    Q^\star_{D,\lambda_t,t}(x_{i,t})
    :=E[Y_{i,t}|X_{t}=x_{i,t}]= E\left[\left.R(X_t,S_{t+1})+\max_{a_{t+1}}Q_{D,\lambda_{t+1},t+1}(  X_t,S_{t+1},a_{t+1})\right|X_t=x_{i,t}\right],
\end{equation}
\end{small}
and
\begin{small}
 \begin{equation}\label{def.Q-1111-condtional-exp-U-dkrr}
    \overline{Q}^\star_{D,\lambda_t,t}(x_{i,t})
    := E[\overline{Y}_{t}|X_t=x_{i,t}]= E\left[\left.R(X_t,S_{t+1})+\max_{a_{t+1}}\overline{Q}_{D,\lambda_{t+1},t+1}(  X_t,S_{t+1},a_{t+1})\right|X_t=x_{i,t}\right].
\end{equation}
\end{small}
The basic idea of our analysis is to develop novel error decomposition strategies to divide $\|Q_{D,\lambda_t,t}-Q_t^*\|_{\mathcal L_t^2}$ into differences among   ${Q}_{D,\lambda_t,t}$, $Q^\oplus_{D,\lambda_t,t}$, $\hat{Q}_{D,\lambda_t,t}$, $Q^\diamond_{D,\lambda_t,t}$ and $\overline{Q}_{D,\lambda_t,t}$, which can be bounded by the difference between $L_{K,D,t}$ and $L_{K,t}$. For this purpose, write
\begin{align}
 \mathcal W_{D,\lambda_t,t}
 &:=
 \left\|(L_{K,t}+\lambda_{t} I)^{-1/2}(L_{K,t}-L_{K,D,t})(L_{K,t}+\lambda_{t} I)^{-1/2}\right\|,\label{def.W}\\
 \mathcal A_{D,\lambda_t,t}
 &:=\left\|\left(L_{K,t}+\lambda_t
    I\right)^{1/2}\left(L_{K,D,t}+\lambda_t
    I\right)^{-1}\left(L_{K,t}+\lambda_t
    I\right)^{1/2}\right\|,\label{Def.A}\\
  \mathcal U_{D,\lambda_t,t,f}
  &:=
  \left\|(L_{K,t}+\lambda_t I)^{-1/2}(L_{K,t}-L_{K,D,t})f\right\|_{K,t},\label{Def.operator-dif}\\
  \mathcal P_{D,\lambda_t,t}
  &:=
  \left\|\left(L_{K,t}+\lambda_t
    I\right)^{-1/2}\left(S_{D,t}^T\mathbf{y}^*_{D,t}-L_{K,D,t}Q^*_t\right)\right\|_{K,t},\label{def.P}\\
   \overline{\mathcal S}_{D,\lambda_t,t}
   &:=\left\|\left(L_{K,t}+\lambda_t
    I\right)^{-1/2}
    \left(S_{D,t}^T\left(\mathbf{y}^*_{D,t}-\overline{\mathbf{y}}_{D,t}\right)-L_{K,D,t}\left(Q^*_t-\overline{Q}^\star_{D,\lambda_t,t}\right)\right)\right\|_{K,t}, \label{Def.conditional-exp-bar}\\
  \mathcal S_{D,\lambda_t,t}
  &:=
  \left\|\left(L_{K,t}+\lambda_t
    I\right)^{-1/2}
    \left(S_{D,t}^T\left(\mathbf{y}^*_{D,t}-\mathbf{y}_{ {D,t}}\right)-L_{K,D,t}\left(Q^*_t-Q^\star_{D,\lambda_t,t}\right)\right)\right\|_{K,t}.\label{Def.conditional-exp}
\end{align}
Our road-map of proofs is to first decompose the generalization error into differences among ${Q}_{D,\lambda_t,t}$, $Q^\oplus_{D,\lambda_t,t}$, $\hat{Q}_{D,\lambda_t,t}$, $Q^\diamond_{D,\lambda_t,t}$, and $\overline{Q}_{D,\lambda_t,t}$, and then quantify the differences via the above terms, and finally derive tight estimates of these terms via some statistical tools.

\subsection{Error decomposition for KRR-DTR}
In this subsection, we present an error decomposition and then derive an oracle inequality for KRR-DTR.
Due to the triangle inequality, we get the following error decomposition directly.

\begin{proposition}\label{Prop:error-dec-KRR}
For any $1\leq t\leq T$, we have
\begin{align}\label{error-dec-KRR}
   &\left\|(L_{K,t}+\lambda_t I)^{1/2}\left(Q_{D,\lambda_t,t}-Q_t^*\right)\right\|_{K,t}
   \leq \left\|(L_{K,t}+\lambda_t I)^{1/2} \left(Q^\diamond_{D,\lambda_t,t}-Q_t^*\right)\right\|_{K,t} \nonumber \\
   +&
   \left\|(L_{K,t}+\lambda_t I)^{1/2}\left(Q^\diamond_{D,\lambda_t,t}-\hat{Q}_{D,\lambda_t,t}\right)\right\|_{K,t}+\left\|(L_{K,t}+\lambda_t I)^{1/2}\left(Q_{D,\lambda_t,t}-\hat{Q}_{D,\lambda_t,t}\right)\right\|_{K,t} .
\end{align}
\end{proposition}

We call the three terms on the right-hand side of \eqref{error-dec-KRR} as the approximation error, sample error, and multi-stage error, respectively. The approximation error, together with some regularity of the optimal Q-functions, describes the approximation capability of the RKHS $\mathcal H_{K,t}$. The sample error quantifies the gap between KRR with the theoretically optimal outputs $Q_t^*$ and KRR with noisy optimal outputs $y_t^*$. The multi-stage error, exclusive to Q-learning, shows the difficulty of Q-learning in circumventing its multi-stage nature. The bounds of approximation error and sample error are standard and can be derived via the recently developed integral operator approach in \citep{blanchard2016convergence,guo2017learning,lin2020distributed}.

\begin{lemma}\label{Lemma:approximation-error-KRR}
Under Assumption 3
with $\frac12\leq r\leq 1$, we have
\begin{equation}\label{approximation-error-KRR}
     \|(L_{K,t}+\lambda_t I)^{1/2} (Q^\diamond_{D,\lambda_t,t}-Q_t^*)\|_{K,t}
      \leq \lambda_t^{r} \mathcal A_{D,\lambda_t,t}  \|h_t\|_{\mathcal L^2_t},\qquad \forall\ t=1,\dots,T.
\end{equation}
\end{lemma}
{\bf Proof.} For any $t=1,\dots,T,$ \eqref{def.without-stage} implies
$$
  Q^\diamond_{D,\lambda_t,t}-Q^*_t
  =-(\left(L_{K,D,t}+\lambda_t
    I\right)^{-1}L_{K,D,t}-I)Q^*_t=-\lambda_t\left(L_{K,D,t}+\lambda_t
    I\right)^{-1} Q^*_t.
$$
Then it follows from \eqref{Def.A}, Assumption 3,
$\|Af\|_K\leq \|A\|\|f\|_K$ for positive operator $A$, and the fact $\left\|L_{K,t}^{1/2}f\right\|_{K,t}=\|f\|_{\mathcal L_t^2}$ that
\vspace{-0.2in}
\begin{align*}
     &\left\|(L_{K,t}\hspace{-0.03in}+\hspace{-0.03in}\lambda_t I)^{1/2}\hspace{-0.01in}(Q^\diamond_{D,\lambda_t,t}\hspace{-0.04in}-\hspace{-0.03in}Q_t^*)\right\|_{K,t}\hspace{-0.05in}\leq\hspace{-0.03in}\lambda_t\hspace{-0.03in}\left\|(\hspace{-0.02in}L_{K,t}\hspace{-0.03in}+\hspace{-0.03in}\lambda_t I)^{1/2}\left(\hspace{-0.02in}L_{K,D,t}\hspace{-0.03in}+\hspace{-0.03in}
      \lambda_t I\right)^{-1}\hspace{-0.05in}(L_{K,t}\hspace{-0.03in}+\hspace{-0.04in}\lambda_t I)^{r-1/2}\right\| \hspace{-0.03in}\left\|L_{K,t}^{1/2}h_t\right\|_{K ,t}\\
     &\leq
      \lambda_t^r\left\|(L_{K,t}+\lambda_t I)^{1/2}\left(L_{K,D,t}+
      \lambda_t I\right)^{-1} (L_{K,t}+\lambda_t I)^{1/2}\right\|\|h_t\|_{\mathcal L_t^2}\leq
      \lambda^r_t\mathcal A_{D,\lambda_t,t}  \|h_t\|_{\mathcal L^2_t}.
\end{align*}
This completes the proof of Lemma \ref{Lemma:approximation-error-KRR}. $\Box$

\begin{lemma}\label{Lemma:sample-error-KRR}
For any $t=1,\dots,T$, it holds that
\begin{equation}\label{sample-error-KRR}
   \left\|(L_{K,t}+\lambda_t I)^{1/2}(Q^\diamond_{D,\lambda_t,t}-\hat{Q}_{D,\lambda_t,t})\right\|_{K,t}
  \leq  \mathcal A_{D,\lambda_t,t} \mathcal P_{D,\lambda_t,t}.
\end{equation}
\end{lemma}
{\bf Proof.} For any $t=1,\dots,T,$ it follows from \eqref{def.without-stage} that
$$
  \hat{Q}_{D,\lambda_t,t}- Q^\diamond_{D,\lambda_t,t}
  =\left(L_{K,D,t}+\lambda_t
    I\right)^{-1} (S_{D,t}^T\mathbf{y}^*_{D,t}-L_{K,D,t}Q^*_t).
$$
Then, \eqref{Def.A}, \eqref{def.P}, and $\|Af\|_K\leq \|A\|\|f\|_K$ for positive operator $A$ yield
\begin{align*}
     &\left\|(L_{K,t}+\lambda_t I)^{1/2}(Q^\diamond_{D,\lambda_t,t}-\hat{Q}_{D,\lambda_t,t})\right\|_{K,t}\\
     \leq&
     \left\|(L_{K,t}+\lambda_{{t}} I)^{1/2}\left(L_{K,D,t}+\lambda_t
    I\right)^{-1}(L_{K,t}+\lambda I)^{1/2}\right\|
     \left\|\left(L_{K,t}+\lambda I\right)^{-1/2} \left(S_{D,t}^T\mathbf{y}^*_{D,t}-L_{K,D,t}Q^*_t\right)\right\|_{K,{t}} \\
    \leq&
     \mathcal A_{D,\lambda_t,t} \mathcal P_{D,\lambda_t,t}.
\end{align*}
This completes the proof of Lemma \ref{Lemma:sample-error-KRR}. $\Box$

The bounding of the multi-stage error is more sophisticated, as collected in the following lemma.

\begin{lemma}\label{Lemma:multi-stage-error-KRR}
Under Assumption 2,
for any $t=1,\dots,T$, it holds that
\begin{align}\label{multi-stage-error}
   & \left\|(L_{K,t}+\lambda_t I)^{1/2}\left(Q_{D,\lambda_t,t}-\hat{Q}_{D,\lambda_t,t}\right)\right\|_{K,t} \notag\\
    \leq &
   \mathcal A_{D,\lambda_t,t} \mathcal S_{D,\lambda,t}+ \mathcal A_{D,\lambda_t,t}\mathcal U_{D,\lambda_t,t,Q^*_t-Q^\star_{D,\lambda_t,t}}
   + \mu^{1/2}\mathcal A_{D,\lambda_t,t}\left\|Q_{D,\lambda_{t+1},t+1}-Q_{t+1}^*\right\|_{\mathcal L_{t+1}^2}.
\end{align}
\end{lemma}
{\bf Proof.} Due to \eqref{operator-KRR} and \eqref{def.without-stage}, we have
\begin{align}\label{KRR-multi-error-dec}
 &  \left\|(\!L_{K,t}\!+\!\lambda_t I)^{1/2}\!\left(\!Q_{D,\lambda_t,t}\!-\!\hat{Q}_{D,\lambda_t,t}\right)\right\|_{K,t}
\!=\!\left\|(\!L_{K,t}\!+\!\lambda_{t} I)^{1/2}\!\left(\!L_{K,D,t}\!+\!\lambda_t
    I\right)^{-1}\!
      S_{D,t}^T\!(\!\mathbf{y}^*_{D,t}\!-\!\mathbf{y}_{D,t})\right\|_{K,t}
 \nonumber\\
    \leq &
    \left\|(L_{K,t}+\lambda_{t} I)^{1/2}\left(L_{K,D,t}+\lambda_t
    I\right)^{-1}
    \left(S_{D,t}^T(\mathbf{y}^*_{D,t}-\mathbf{y}_{D,t})-L_{K,D,t}(Q^*_t-Q^\star_{D,\lambda_t,t})\right)\right\|_{K,t} \nonumber \\
    +&
    \left\|(L_{K,t}+\lambda_{t} I)^{1/2}\left(L_{K,D,t}+\lambda_t
    I\right)^{-1}(L_{K,D,t}-L_{K,t})(Q^*_t-Q^\star_{D,\lambda_t,t})\right\|_{K,t}   \nonumber\\
    +&
    \left\|(L_{K,t}+\lambda_{t} I)^{1/2}\left(L_{K,D,t}+\lambda_t
    I\right)^{-1}L_{K,t}(Q^*_t-Q^\star_{D,\lambda_t,t})\right\|_{K,t}.
\end{align}
It follows from \eqref{Def.A} and \eqref{Def.conditional-exp} that
\begin{equation}\label{krr-multi-first-term}
    \left\|(\!L_{K,t}\!+\!\lambda_{t} I)^{1/2}\!\left(\!L_{K,D,t}\!+\!\lambda_t
    I\right)^{-1}\!
    \left(\!S_{D,t}^T(\mathbf{y}^*_{D,t}\!-\!\mathbf{y}_{D,t})\!-\!L_{K,D,t}(Q^*_t\!-\!Q^\star_{D,\lambda_t,t})\right)\right\|_{K,t}\!\leq \!\mathcal A_{D,\lambda_t,t} \mathcal S_{D,\lambda_t,t}.
\end{equation}
Furthermore, \eqref{Def.A} and \eqref{Def.operator-dif} yield
\begin{equation}\label{krr-multi-second-term}
    \left\|(L_{K,t}+\lambda_{t} I)^{1/2}\left(L_{K,D,t}+\lambda_t
    I\right)^{-1}(L_{K,D,t}-L_{K,t})(Q^*_t-Q^\star_{D,\lambda_t,t})\right\|_{K,t}
    \leq
    \mathcal A_{D,\lambda_t,t}\mathcal U_{D,\lambda_t,t,Q^*_t-Q^\star_{D,\lambda_t,t}}.
\end{equation}
To bound the last term in \eqref{KRR-multi-error-dec}, we notice
\begin{equation*}
\left\|(L_{K,t}+\lambda_{t} I)^{1/2}\left(L_{K,D,t}+\lambda_t
    I\right)^{-1}L_{K,t}(Q^*_t-Q^\star_{D,\lambda_t,t})\right\|_{K,t}
    \leq
    \mathcal A_{D,\lambda_t,t}\left\|Q^*_t-Q^\star_{D,\lambda_t,t}\right\|_{\mathcal L_t^2}.
\end{equation*}
But Assumption 2, (4), (17), and (29) show
\begin{align*}
  & \left\|Q^*_t-Q^\star_{D,\lambda_t,t}\right\|_{\mathcal L_t^2}^2
   =
   E\left[\left.\left(Q^\star_{D,\lambda_t,t}(X_{t})-Q^*_t(X_{t}) \right)^2 \right|D\right]\\
   = &
  E\left[\left.\left(\max_{a_{t+1}}Q_{D,\lambda_{t+1},t+1}(  S_{1:t+1},
     A_{1:t},a_{t+1})-\max_{a_{t+1}}Q_{t+1}^*(  S_{1:t+1},
     A_{1:t},a_{t+1})\right)^2 \right|D\right]\\
     \leq &
     E\left[\left.\max_{a_{t+1}}\left(Q_{D,\lambda_{t+1},t+1}(  S_{1:t+1},
     A_{1:t},a_{t+1})-Q_{t+1}^*(  S_{1:t+1},
     A_{1:t},a_{t+1})\right)^2 \right|D\right]
\end{align*}
\begin{align*}
     \leq &
     E\left[\left.\mu\sum_{a\in\mathcal A_{t+1}}\left(Q_{D,\lambda_{t+1},t+1}(  S_{1:t+1},
     A_{1:t},a)-Q_{t+1}^*(  S_{1:t+1},
     A_{1:t},a)\right)^2p_t(a|S_{1:t+1},A_{1:t})\right|D\right]\\
     = &
     \mu E\left[\left.\left(Q_{D,\lambda_{t+1},t+1}-Q_{t+1}^*\right)^2\right|D\right]
     =\mu\left\|Q_{D,\lambda_{t+1},t+1}-Q_{t+1}^*\right\|_{\mathcal L_{t+1}^2}^2.
\end{align*}
Therefore,
\begin{equation}\label{KRR-multi-third-term}
    \left\|(L_{K,t}\hspace{-0.03in}+\hspace{-0.03in}\lambda_{t} I)^{1/2}\left(L_{K,D,t}\hspace{-0.03in}+\hspace{-0.03in}\lambda_t
    I\right)^{-1}\hspace{-0.03in}L_{K,t}(Q^*_t\hspace{-0.03in}-\hspace{-0.03in}Q^\star_{D,\lambda_t,t})\right\|_{K,t}
    \hspace{-0.03in}\leq \hspace{-0.03in}\mu^{1/2}\mathcal A_{D,\lambda_t,t}\hspace{-0.03in}\left\|Q_{D,\lambda_{t+1},t+1}\hspace{-0.03in}-\hspace{-0.03in}Q_{t+1}^*\right\|_{\mathcal L_{t+1}^2}.
\end{equation}
Plugging \eqref{krr-multi-first-term}, \eqref{krr-multi-second-term}, \eqref{KRR-multi-third-term} into \eqref{KRR-multi-error-dec},
we obtain \eqref{multi-stage-error} directly.
This finishes the proof of Lemma \ref{Lemma:multi-stage-error-KRR}. $\Box$

Using Proposition \ref{Prop:error-dec-KRR}, Lemma \ref{Lemma:approximation-error-KRR}, Lemma \ref{Lemma:sample-error-KRR} and Lemma \ref{Lemma:multi-stage-error-KRR}, we can derive the following oracle inequality for KRR-DTR.

\begin{proposition}\label{Proposition:Oracle-KRR}
Under Assumption 2 and Assumption 3 with $\frac12\leq r\leq 1$, we have
\begin{align*}
    &\left\|(L_{K,t}+\lambda_t I)^{1/2}(Q_{D,\lambda_t,t}-Q_t^*)\right\|_{K,t} \\
    \leq&
    \sum_{\ell=t}^T\left(\mu^{\frac12}\mathcal A_{D,\lambda_\ell,\ell}\right)^{\ell-t}\mathcal A_{D,\lambda_\ell,\ell}  \left(\lambda_t^{r}\|h_t\|_{\mathcal L^2_t}
   +\mathcal P_{D,\lambda_\ell,\ell}
   +\mathcal S_{D,\lambda_\ell,\ell}+ \mathcal U_{D,\lambda_\ell,\ell,Q^*_\ell-Q^\star_{D,\lambda_\ell,\ell}}\right).
\end{align*}
\end{proposition}

{\bf Proof.} Inserting \eqref{approximation-error-KRR}, \eqref{sample-error-KRR} and \eqref{multi-stage-error} into \eqref{error-dec-KRR}, we obtain for any $t=1,2,\dots,T$,
\vspace{-0.2in}
\begin{align*}
&  \left\|(L_{K,t}+\lambda_t I)^{1/2}(Q_{D,\lambda_t,t}-Q_t^*)\right\|_{K,t}\\
 \leq&
   \mathcal A_{D,\lambda_t,t} \!\left(\!\lambda_t^{r}\|h_t\|_{\mathcal L^2_t}
   \!+\!\mathcal P_{D,\lambda_t,t}
   \!+\!\mathcal S_{D,\lambda_t,t}\!+\!\mathcal U_{D,\lambda_t,t,Q^*_t-Q^\star_{D,\lambda_t,t}}\right)
   \!+\!\mu^{\frac12}\mathcal A_{D,\lambda_t,t}\|Q_{D,\lambda_{t+1},t+1}\!-\!Q_{t+1}^*\|_{\mathcal L_{t+1}^2}.
\end{align*}
It then follows from $Q_{D,\lambda_{T+1},T+1}=Q_{T+1}^*=0$ that
\begin{align*}
     &\left\|(L_{K,t}+\lambda_t I)^{1/2}(Q_{D,\lambda_t,t}-Q_t^*)\right\|_{K,t} \\
    \leq&
    \sum_{\ell=t}^T\!\left(\mu^{\frac12}\mathcal A_{D,\lambda_\ell,\ell}\right)^{\ell-t}\mathcal A_{D,\lambda_\ell,\ell}   \left(\lambda_t^{r}\|h_t\|_{\mathcal L^2_t}
   +\mathcal P_{D,\lambda_\ell,\ell}
    + \mathcal S_{D,\lambda_\ell,\ell}+ \mathcal U_{D,\lambda_\ell,\ell,Q^*_\ell-Q^\star_{D,\lambda_\ell,\ell}}\right).
\end{align*}
The proof of  Proposition \ref{Proposition:Oracle-KRR} is completed. $\Box$

\subsection{Error decomposition for DKRR-DTR}

The generalization error of DKRR-DTR also requires the following error decomposition.
\begin{proposition}\label{Prop:error-decom-DKRR}
For any $1\leq t\leq T$, it holds that
\begin{align*}
&\left\|(L_{K,t}+\lambda_t I)^{1/2}\left(\overline{Q}_{D,{\lambda}_t,t}-Q_t^*\right)\right\|_{K,t}\\
    \leq&
     \sum_{j=1}^m\frac{|D_j|}{|D|}\mathcal W_{D_j,\lambda_t,t}\left\|(L_{K,t}+\lambda_tI)^{1/2}(Q_{D_j,\lambda_{t},t}-Q_t^*)\right\|_{K,t}
    +     \mathcal U_{D,\lambda_t,t,Q_t^*}
    +    \left\|Q^\oplus_{D,\lambda_t,t}-Q_t^*    \right\|_{\mathcal L_t^2}.
\end{align*}
\end{proposition}

{\bf Proof.}  Similar to \eqref{operator-KRR}, it follows from
(13) and (14) that
$$
    \overline{Q}_{D,{\lambda}_t,t}=\sum_{j=1}^m\frac{|D_j|}{|D|}(L_{K,D_j,t}+\lambda_{t}I)^{-1}S_{D_j,t}^T\overline{\bf y}_{j,t},
$$
where $\overline{\bf y}_{j,t}:=\left(\overline{y}_{1,j,t},\dots, \overline{y}_{|D_j|,j,t}\right)^T$. Then
\begin{align*}
   &&\left\|(L_{K,t}\hspace{-0.02in}+\hspace{-0.02in}\lambda_t I)^{1/2}(\overline{Q}_{D,{\lambda}_t,t}\hspace{-0.02in}-\hspace{-0.02in}Q_t^*)\right\|_{K,t} \hspace{-0.02in}\leq \hspace{-0.02in}\left\|(L_{K,t}\hspace{-0.02in}+\hspace{-0.02in}\lambda_t I)^{1/2} {\left( \sum_{j=1}^m\frac{|D_j|}{|D|}(L_{K,t}\hspace{-0.02in}+\hspace{-0.02in}\lambda_{t}I)^{-1}S_{D_j,t}^T\overline{\bf y}_{j,t}\hspace{-0.02in} -\hspace{-0.02in} Q_t^*\right) }    \right\|_{K,t} \\
   &&+
    \left\|(L_{K,t}+\lambda_t I)^{1/2}\sum_{j=1}^m\frac{|D_j|}{|D|}\left((L_{K,D_j,t}+\lambda_{t}I)^{-1}-(L_{K,t}+\lambda_{t}I)^{-1}\right)S_{D_j,t}^T\overline{\bf y}_{j,t} \right\|_{K,t}.
\end{align*}
But
\begin{align*}
   & \left((L_{K,D_j,t}+\lambda_{t}I)^{-1}-(L_{K,t}+\lambda_{t}I)^{-1}\right) S_{D_j,t}^T\overline{\bf y}_{j,t}
   =
   (L_{K,t}+\lambda_{t} I)^{-1}(L_{K,t}-L_{K,D_j,t})Q_{D_j,\lambda_{t},t}\\
   =&
   (L_{K,t}+\lambda_{t} I)^{-1}(L_{K,t}-L_{K,D_j,t})(Q_{D_j,\lambda_{t},t}-Q_t^*)
   +
   (L_{K,t}+\lambda_{t} I)^{-1}(L_{K,t}-L_{K,D_j,t})Q_t^*,
\end{align*}
\begin{equation*}
  \sum_{j=1}^m\frac{|D_j|}{|D|}(L_{K,t}+\lambda_{t}I)^{-1}S_{D_j,t}^T\overline{\bf y}_{j,t}
  = (L_{K,t}+\lambda_{t}I)^{-1}S_{D,t}^T \overline{\bf y}_{t},
\end{equation*}
and
$$
   \sum_{j=1}^m\frac{|D_j|}{|D|}(L_{K,t}+\lambda_{t} I)^{-1}(L_{K,t}-L_{K,D_j,t})Q_t^*
   =
   (L_{K,t}+\lambda_{t} I)^{-1}(L_{K,t}-L_{K,D,t})Q_t^*.
$$
Therefore, Jensen's inequality together with \eqref{Def.Qoplus-new} yields
\begin{align*}
   &\left\|\!(\!L_{K,t}\!+\!\lambda_t I)^{1/2}(\overline{Q}_{D,{\lambda}_t,t}\!-\!Q_t^*)\!\right\|_{K,t}
   \!\leq\!
    \left\|\sum_{j=1}^m\!\frac{|D_j|}{|D|}(L_{K,t}\!+\!\lambda_{t} I)^{-1/2}(\!L_{K,t}\!-\!L_{K,D_j,t}\!)\!(\!Q_{D_j,\lambda_{t},t}\!-\!Q_t^*) \right\|_{K,t}\\
    +&
    \left\| (L_{K,t}+\lambda_{t} I)^{-1/2}(L_{K,t}-L_{K,D,t})Q_t^* \right\|_{K,t}
    +{
    \left\|(L_{K,t}+\lambda_t I)^{1/2}(Q^\oplus_{D,\lambda_t,t}-Q_t^*)    \right\|_{K,t} }\\
    \leq&\!
    \sum_{j=1}^m\!\frac{|D_j|}{|D|}\mathcal W_{D_j,\lambda_t,t}\|(L_{K,t}+\lambda_tI )^{1/2}(Q_{D_j,\lambda_{t},t}-Q_t^*)\|_{K,t}
    + \mathcal U_{D,\lambda_t,t,Q_t^*}\\
    & \qquad +
    \left\|\!(\!L_{K,t}\!+\!\lambda_t I)^{1/2}(Q^\oplus_{D,\lambda_t,t}\!-\!Q_t^* ) \right\|_{K,t}.
\end{align*}
This completes the proof of Proposition \ref{Prop:error-decom-DKRR}. $\Box$

Our following lemma presents an upper bound of $\left\|(L_{K,t}+\lambda_t I)^{1/2}(Q^\oplus_{D,\lambda_t,t}-Q_t^* ) \right\|_{K,t}$.

\begin{lemma}\label{Lemma:bound-DKRR-1-important}
Under Assumption 2 and Assumption 3
with $\frac12\leq r\leq 1$,  for $\forall$ $t=1,\dots,T$, there holds
\begin{align*}
   \left\|(L_{K,t}+\lambda_t I)^{1/2}(Q^\oplus_{D,\lambda_t,t}-Q_t^*)   \right\|_{K,t}
   \leq &
  \mathcal A_{D,\lambda_t,t}  \left(\lambda_t^{r}  \|h_t\|_{\mathcal L^2_t}+  \mathcal P_{D,\lambda_t,t}+   \overline{\mathcal S}_{D,\lambda_t,t}
    +
     \mathcal U_{D,\lambda_t,t,Q^*_t-\overline{Q}^\star_{D,\lambda_t,t}}\right)\\
    & +
    \mu^{1/2}\mathcal A_{D,\lambda_t,t}\|\overline{Q}_{D,\lambda_{t+1},t+1}-Q_{t+1}^*\|_{\mathcal L_{t+1}^2}.
\end{align*}
\end{lemma}
{\bf Proof.} The proof of the lemma is almost the same as that in bounding $\|(L_{K,t}+\lambda_t I)^{1/2}(Q_{D,\lambda_t,t}-Q_t^*)\|_{K,t}$ in the previous subsection. For the sake of completeness, we sketch the proof. Due to the triangle inequality, it holds that
\begin{align*}
  &\left\|(L_{K,t}+\lambda_t I)^{1/2}(Q^\oplus_{D,\lambda_t,t}-Q_t^*)    \right\|_{K,t}
   \leq \|(L_{K,t}+\lambda_t I)^{1/2}( Q^\diamond_{D,\lambda_t,t}-Q_t^*)\|_{K,t}\\
   +&
   \|(L_{K,t}+\lambda_t I)^{1/2}(Q^\diamond_{D,\lambda_t,t}-\hat{Q}_{D,\lambda_t,t})\|_{K,t}+\|(L_{K,t}+\lambda_t I)^{1/2}(Q^\oplus_{D,\lambda_t,t}-\hat{Q}_{D,\lambda_t,t})\|_{K,t}.
\end{align*}
Then, it follows from Lemma \ref{Lemma:approximation-error-KRR} and Lemma \ref{Lemma:sample-error-KRR} that
\begin{align*}
   \left\|(L_{K,t}+\lambda_t I)^{1/2}(Q^\oplus_{D,\lambda_t,t}-Q_t^*) \right\|_{K,t}
  \leq & \lambda_t^{r} \mathcal A_{D,\lambda_t,t}  \|h_t\|_{\mathcal L^2_t}+\mathcal A_{D,\lambda_t,t} \mathcal P_{D,\lambda_t,t}\\
  & +\|(L_{K,t}+\lambda_t I)^{1/2}(Q^\oplus_{D,\lambda_t,t}-\hat{Q}_{D,\lambda_t,t})\|_{K,t}.
\end{align*}
Similar as the approach in proving Lemma \ref{Lemma:multi-stage-error-KRR}, we obtain from  \eqref{def.without-stage}  and  \eqref{Def.Qoplus-new} that
\begin{align*}
 &   \left\|(L_{K,t}\!+\!\lambda_t I)^{1/2}(Q^\oplus_{D,\lambda_t,t}\!-\!\hat{Q}_{D,\lambda_t,t})\right\|_{K,t} \leq
    \left\|(L_{K,t}\!+\!\lambda I)^{1/2}\left(L_{K,D,t}\!+\!\lambda_t
    I\right)^{-1}L_{K,t}(Q^*_t\!-\!\overline{Q}^\star_{D,\lambda_t,t})\right\|_{K,t}
     \nonumber\\
    &+
    \left\|(L_{K,t}+\lambda I)^{1/2}\left(L_{K,D,t}+\lambda_t
    I\right)^{-1}(L_{K,D,t}-L_{K,t})(Q^*_t-\overline{Q}^\star_{D,\lambda_t,t})\right\|_{K,t}   \nonumber\\
    &+
    \left\|(L_{K,t}+\lambda I)^{1/2}\left(L_{K,D,t}+\lambda_t
    I\right)^{-1}
    \left(S_{D,t}^T(\mathbf{y}^*_{D,t}-\overline{\mathbf{y}}_{D,t})-L_{K,D,t}(Q^*_t-\overline{Q}^\star_{D,\lambda_t,t})\right)\right\|_{K,t} \nonumber\\
    &\leq
    \mathcal A_{D,\lambda_t,t} \overline{\mathcal S}_{D,\lambda_t,t}
    +
    \mathcal A_{D,\lambda_t,t}\mathcal U_{D,\lambda_t,t,Q^*_t-\overline{Q}^\star_{D,\lambda_t,t}}
    +
    \mu^{1/2}\mathcal A_{D,\lambda_t,t}\left\|\overline{Q}_{D,\lambda_{t+1},t+1}-Q_{t+1}^*\right\|_{\mathcal L_{t+1}^2}.
\end{align*}
Therefore,
\begin{align*}
   \left\|(L_{K,t}+\lambda_t I)^{1/2}(Q^\oplus_{D,\lambda_t,t}-Q_t^* )   \right\|_{ {K,t}} \leq  &
  \mathcal A_{D,\lambda_t,t}  \left(\lambda_t^{r}  \|h_t\|_{\mathcal L^2_t}+  \mathcal P_{D,\lambda_t,t}+   \overline{\mathcal S}_{D,\lambda_t,t}
    +
     \mathcal U_{D,\lambda_t,t,Q^*_t-\overline{Q}^\star_{D,\lambda_t,t}}\right)\\
     &   +
    \mu^{1/2}\mathcal A_{D,\lambda_t,t}\left\|\overline{Q}_{D,\lambda_{t+1},t+1}-Q_{t+1}^*\right\|_{\mathcal L_{t+1}^2}.
\end{align*}
This completes the proof of Lemma \ref{Lemma:bound-DKRR-1-important}. $\Box$

We also bound $\|(L_{K,t}+\lambda I)^{1/2}(Q_{D_j,\lambda_t,t}-Q_t^*)\|_{K,t}$ in the following lemma.

\begin{lemma}\label{Lemma:bound-DKRR-2-important}
Under Assumption 2 and Assumption 3
with $\frac12\leq r\leq 1$, for $\forall$ $t=1,\dots,T$, there holds
\begin{align*}
  \left\|\!(\!L_{K,t}\!+\!\lambda_t I)^{1/2}(Q_{D_j,\lambda_t,t}\!-\!Q_t^* )  \right\|_{K,t}
   \leq&
    \mathcal A_{D_j,\lambda_t,t} \! \left(\!\lambda_t^{r} \|h_t\|_{\mathcal L^2_t}\!+\! \mathcal P_{D_j,\lambda_t,t}
   \!+\!
   \overline{\mathcal S}_{D_j,\lambda_t,t}
    \!+\!
    \mathcal U_{D_j,\lambda_t,t,Q^*_t-\overline{Q}^\star_{D_j,\lambda_t,t}}\!\right)\\
    &+
     \mu^{1/2}\mathcal A_{D_j,\lambda_t,t}\|\overline{Q}_{D,\lambda_{t+1},t+1}-Q_{t+1}^*\|_{\mathcal L_{t+1}^2}.
\end{align*}
\end{lemma}

{\bf Proof.} It follows from  Lemma \ref{Lemma:approximation-error-KRR} and Lemma \ref{Lemma:sample-error-KRR} with $D=D_j$ that
\begin{align*}
  &\left\|(L_{K,t}+\lambda_t I)^{1/2}(Q_{D_j,\lambda_t,t}-Q_t^*)   \right\|_{K,t}
   \leq \left\|(L_{K,t}+\lambda_t I)^{1/2} \left(Q^\diamond_{D_j,\lambda_t,t}-Q_t^*\right)\right\|_{K,t} \\
   &+
   \left\|(L_{K,t}+\lambda_t I)^{1/2}\left(Q^\diamond_{D_j,\lambda_t,t}-\hat{Q}_{D_j,\lambda_t,t}\right)\right\|_{K,t} +\left\|(L_{K,t}+\lambda_t I)^{1/2}(Q_{D_j,\lambda_t,t}-\hat{Q}_{D_j,\lambda_t,t})\right\|_{K,t}\\
   &\leq
   \lambda_t^{r} \mathcal A_{D_j,\lambda_t,t}  \|h_t\|_{\mathcal L^2_t}+\mathcal A_{D_j,\lambda_t,t} \mathcal P_{D_j,\lambda_t,t}
   +
   \left\|(L_{K,t}+\lambda_t I)^{1/2}(Q_{D_j,\lambda_t,t}-\hat{Q}_{D_j,\lambda_t,t})\right\|_{K,t}.
\end{align*}
But for any $t=1,\dots,T$, we have from (13),
\eqref{def.without-stage-DKRR}, and the same method in proving Lemma \ref{Lemma:multi-stage-error-KRR} that
\begin{align*}
 &  \left\|(\!L_{K,t}\!+\!\lambda_t I)^{1/2}(\!Q_{D_j,\lambda_t,t}\!-\!\hat{Q}_{D_j,\lambda_t,t})\right\|_{K,t}  \!\leq\!
    \left\|(\!L_{K,t}\!+\!\lambda I)^{1/2}\left(\!L_{K,D_j,t}\!+\!\lambda_t
    I\right)^{-1}\!L_{K,t}(Q^*_t\!-\!\overline{Q}^\star_{D_j,\lambda_t,t})\right\|_{K,t}\nonumber\\
    &+
    \left\|(L_{K,t}+\lambda I)^{1/2}\left(L_{K,D_j,t}+\lambda_t
    I\right)^{-1}
    \left(S_{D_j,t}^T\left(\mathbf{y}^*_{D_j,t}-\overline{\mathbf{y}}_{ D_j,t}\right)-L_{K,D,t}\left(Q^*_t-\overline{Q}^\star_{D_j,\lambda_t,t}\right)\right)\right\|_{K,t} \nonumber \\
    &+
    \left\|(L_{K,t}+\lambda I)^{1/2}\left(L_{K,D_j,t}+\lambda_t
    I\right)^{-1}(L_{K,D_j,t}-L_{K,t})\left(Q^*_t-\overline{Q}^\star_{D_j,\lambda_t,t}\right)\right\|_{K,t}   \nonumber\\
    &\leq
    \mathcal A_{D_j,\lambda_t,t} \overline{\mathcal S}_{D_j,\lambda_t,t}
    +
    \mathcal A_{D_j,\lambda_t,t}\mathcal U_{D_j,\lambda_t,t,Q^*_t-\overline{Q}^\star_{D_j,\lambda_t,t}}
    +
    \mu^{1/2}\mathcal A_{D_j,\lambda_t,t}\left\|\overline{Q}_{D,\lambda_{t+1},t+1}-Q_{t+1}^*\right\|_{\mathcal L_{t+1}^2}.
\end{align*}
Therefore,
\begin{align*}
  \left\|(L_{K,t}\!+\!\lambda_t I)^{1/2}(Q_{D_j,\lambda_t,t}\!-\!Q_t^* )  \right\|_{K,t}
   \leq &
    \mathcal A_{D_j,\lambda_t,t} \! \left( \!\lambda_t^{r} \|h_t\|_{\mathcal L^2_t}\!+ \!\mathcal P_{D_j,\lambda_t,t}
   \!+\!
   \overline{\mathcal S}_{D_j,\lambda_t,t}
   \! +\!
    \mathcal U_{D_j,\lambda_t,t,Q^*_t-\overline{Q}^\star_{D_j,\lambda_t,t}}\!\right)\\
    &+
     \mu^{-1/2}\mathcal A_{D_j,\lambda_t,t}\|\overline{Q}_{D,\lambda_{t+1},t+1}-Q_{t+1}^*\|_{\mathcal L_{t+1}^2}.
\end{align*}
This completes the proof of Lemma \ref{Lemma:bound-DKRR-2-important}. $\Box$

Based on Lemma \ref{Lemma:bound-DKRR-1-important}, Lemma  \ref{Lemma:bound-DKRR-2-important},  and Proposition \ref{Prop:error-decom-DKRR}, we can derive the following oracle inequality.

\begin{proposition}\label{Prop:DKRR-Oracle}
Under
Assumption 2 and Assumption 3 with $\frac12\leq r\leq 1$, it holds that
\begin{align*}
    & \left\|(L_{K,t}+\lambda_t I)^{1/2}\left(\overline{Q}_{D,{\lambda}_t,t}-Q_t^*\right)\right\|_{K,t}\\
    \leq&
     \sum_{\ell=t}^T\prod_{k=t+1}^\ell \left(\left(\sum_{j=1}^m\frac{|D_j|}{|D|}\mathcal W_{D_j,\lambda_k,k}  \mathcal A_{D_j,\lambda_k,k}
    +
    \mathcal A_{D,\lambda_k,k}\right) \mu^{1/2}\right)\\
    & \times
    \left(\sum_{j=1}^m\frac{|D_j|}{|D|}\mathcal W_{D_j,\lambda_\ell,\ell}
      \mathcal A_{D_j,\lambda_\ell,\ell}  \left(\lambda_\ell^{r} \|h_\ell\|_{\mathcal L^2_\ell}+ \mathcal P_{D_j,\lambda_\ell,\ell}
   +
   \overline{\mathcal S}_{D_j,\lambda_\ell,\ell}
    +
    \mathcal U_{D_j,\lambda_\ell,\ell,Q^*_\ell-\overline{Q}^\star_{D_j,\lambda_\ell,\ell}} \right)\right.\\
    & +
    \left.\mathcal A_{D,\lambda_\ell,\ell}  \left(\lambda_\ell^{r}  \|h_\ell\|_{\mathcal L^2_\ell}+  \mathcal P_{D,\lambda_\ell,\ell}+   \overline{\mathcal S}_{D,\lambda_\ell,\ell}
    +\mathcal U_{D,\lambda_\ell,\ell,Q^*_\ell-\overline{Q}^\star_{D,\lambda_\ell,\ell}}\right)
     +
    \mathcal U_{D,\lambda_\ell,\ell,Q_\ell^*}\right).
\end{align*}
\end{proposition}

{\bf Proof.} Inserting Lemma \ref{Lemma:bound-DKRR-1-important} and Lemma  \ref{Lemma:bound-DKRR-2-important} into Proposition \ref{Prop:error-decom-DKRR}, we obtain
\begin{align*}
     &\left\|(L_{K,t}+\lambda_t I)^{1/2}\left(\overline{Q}_{D,{\lambda}_t,t}-Q_t^*\right)\right\|_{K,t}\\
    \leq&
     \sum_{j=1}^m\frac{|D_j|}{|D|}\mathcal W_{D_j,\lambda_t,t}
      \mathcal A_{D_j,\lambda_t,t}  \left(\lambda_t^{r} \|h_t\|_{\mathcal L^2_t}+ \mathcal P_{D_j,\lambda_t,t}
   +
   \overline{\mathcal S}_{D_j,\lambda_t,t}
    +
    \mathcal U_{D_j,\lambda_t,t,Q^*_t-\overline{Q}^\star_{D_j,\lambda_t,t}}\right)\\
    &+
     \mathcal A_{D,\lambda_t,t}  \left(\lambda_t^{r}  \|h_t\|_{\mathcal L^2_t}+  \mathcal P_{D,\lambda_t,t}+   \overline{\mathcal S}_{D,\lambda_t,t}
    +
     \mathcal U_{D,\lambda_t,t,Q^*_t-\overline{Q}^\star_{D,\lambda_t,t}}\right)
     +
    \mathcal U_{D,\lambda_t,t,Q_t^*}\\
    &+
     \left(\sum_{j=1}^m\frac{|D_j|}{|D|}\mathcal W_{D_j,\lambda_t,t}  \mathcal A_{D_j,\lambda_t,t}
    +
    \mathcal A_{D,\lambda_t,t}\right) \mu^{1/2}\left\|\overline{Q}_{D,\lambda_{t+1},t+1}-Q_{t+1}^*\right\|_{\mathcal L_{t+1}^2}\\
    \leq&
    \sum_{\ell=t}^T\prod_{k=t+1}^\ell \left(\left(\sum_{j=1}^m\frac{|D_j|}{|D|}\mathcal W_{D_j,\lambda_k,k}  \mathcal A_{D_j,\lambda_k,k}
    +
    \mathcal A_{D,\lambda_k,k}\right) \mu^{1/2}\right) \\
    &\times
    \left(\sum_{j=1}^m\frac{|D_j|}{|D|}\mathcal W_{D_j,\lambda_\ell,\ell}
      \mathcal A_{D_j,\lambda_\ell,\ell}  \left(\lambda_\ell^{r} \|h_\ell\|_{\mathcal L^2_\ell}+ \mathcal P_{D_j,\lambda_\ell,\ell}
   +
   \overline{\mathcal S}_{D_j,\lambda_\ell,\ell}
    +
    \mathcal U_{D_j,\lambda_\ell,\ell,Q^*_\ell-\overline{Q}^\star_{D_j,\lambda_\ell,\ell}} \right)\right.\\
    &+
    \left.\mathcal A_{D,\lambda_\ell,\ell}  \left(\lambda_\ell^{r}  \|h_\ell\|_{\mathcal L^2_\ell}+  \mathcal P_{D,\lambda_\ell,\ell}+   \overline{\mathcal S}_{D,\lambda_\ell,\ell}
    +\mathcal U_{D,\lambda_\ell,\ell,Q^*_\ell-\overline{Q}^\star_{D,\lambda_\ell,\ell}} \right)
     +
    \mathcal U_{D,\lambda_\ell,\ell,Q_\ell^*}\right).
\end{align*}
This completes the proof of Proposition \ref{Prop:DKRR-Oracle}. $\Box$

\subsection{Bounds of operator differences}
In this part, we aim at bounding the operator differences. The first lemma focuses on bounding $\mathcal W_{D,\lambda_t,t}$, which can be found in \citep[Lemma 6]{lin2020distributed}.
\begin{lemma}\label{Lemma:WD}
Let $0<\delta\leq 1$. If Assumption 1 holds and
$0<\lambda\leq1$, then  with confidence $1-\delta,$ it holds that
$$
         \mathcal W_{D,\lambda_t,t}
         \leq C_1^*\mathcal B_{D,\lambda_t,t}\log\frac4\delta,\qquad t=1,\dots,T,
$$
where $C_1^*:=\max\{(\kappa^2+1)/3,2\sqrt{\kappa^2+1}\}$ and
\begin{equation}\label{Def.B}
     \mathcal B_{D,\lambda_t,t}:=
           \frac{1+\log (1+\mathcal
      N_t(\lambda_t))}{\lambda|D|}+ \sqrt{\frac{1+\log (1+\mathcal
      N_t(\lambda_t))}{\lambda|D|}}.
\end{equation}
\end{lemma}

The second one aims at bounding $\mathcal A_{D,\lambda_t,t}$, which can be found in \cite[Lemma 7]{lin2020distributed}.

\begin{lemma}\label{Lemma:AD}
If Assumption 1
holds and   $0<\lambda\leq 1$, then for
  $\delta\geq 4\exp\{-1/(2C_1^*\mathcal B_{D,\lambda_t,t})\}$ , with
confidence $1-\delta,$ it holds that
$\mathcal A_{D,\lambda_t,t}\leq 2$ for $t=1,\dots,T$.
\end{lemma}

The bound of $\mathcal U_{D,\lambda_t,t,f}$, provided in \cite[Lemma 18]{lin2017distributed}, is shown in the following lemma.

\begin{lemma}\label{Lemma:UD}
Under Assumption 1,
for any $0< \delta <1$ and $\|f\|_{L^\infty}<\infty$, with confidence at least $1-\delta$, it holds that
\begin{equation*}
  \mathcal U_{D,\lambda_t,t,f} \leq \frac{2 \|f\|_{L^\infty} \log \bigl(2/\delta\bigr) }{\sqrt{|D|}} \left\{\frac{\kappa }{\sqrt{|D|\lambda_t}} +\sqrt{{\mathcal N}(\lambda_t)}\right\},\qquad t=1,\dots,T.
\end{equation*}
\end{lemma}

To bound $\mathcal P_{D,\lambda_t,t}$, $\mathcal S_{D,\lambda_t,t}$ and $\overline{\mathcal S}_{D,\lambda_t,t}$, we should introduce a Hilbert valued Berstein inequality established in \citep{pinelis1994optimum}.

\begin{lemma}\label{Lemma:Pinelislemma}
For a random variable $\xi$ on $({\mathcal Z}, \rho)$ with values in
a separable Hilbert space $(H, \|\cdot\|)$ satisfying $\|\xi\| \leq
\bar{M} <\infty$ almost surely, and a random sample
$\{z_i\}_{i=1}^s$ independent drawn according to $\rho$, there holds
with confidence $1-\widetilde{\delta}$ that
$$
 \left\|\frac1s\sum_{i=1}^s (\xi (z_i) -
 E[\xi])\right\| \leq {2 \bar{M} \log
 \bigl(2/\widetilde{\delta}\bigr) \over s}  + \sqrt{{2 E [\|\xi\|^2]
 \log \bigl(2/\widetilde{\delta}\bigr) \over s}}.
$$
\end{lemma}

With the help of Lemma \ref{Lemma:Pinelislemma}, we can derive the following concentration inequality, which is crucial to bound $\mathcal P_{D,\lambda_t,t}, \mathcal S_{D,\lambda_t,t}$ and $\overline{\mathcal S}_{D,\lambda_t,t}$.

\begin{lemma}\label{Lemma:Concentration-1}
Let $0<\delta<1$, ${\bf y}_D=\left(y_1,\dots,y_{|D|}\right)^T\in \mathbb R^{|D|}$  with $|y_i|\leq\tilde{M}$ almost surely, and $g(x_i)=E[Y_i|x_i]$.
Under Assumption 1,
with confidence $1-\delta$, it holds that
$$
   \|(L_{K,t}+\lambda_t I)^{-1/2}(S^T_{D,t}{\bf y}_D-L_{K,D}g)\|_{K,t}  \leq
  \frac{ 2\sqrt{2}\tilde{M}\log(2/\delta)}{\sqrt{|D|}}\left(
\frac{\sqrt{2}}{  \sqrt{|D|\lambda_t}}+\sqrt{\mathcal N_t(\lambda_t)}\right).
$$
\end{lemma}

{\bf Proof.}
Denote $z_{i}=(x_{i},y_{i})$ and $\eta_t(z_{i})=(L_{K,t}+\lambda_t I)^{-1/2}(y_{i}-g(x_{i}))K_{x_{i}}\in\mathcal H_{K,t}$.
Then, we have
$
 E[\eta_t(z_{i})]=0$ and
\begin{align*}
    \|\eta_t(z_{i})\|_{K,t}^2
    &=
    \langle (L_{K,t}+\lambda_t I)^{-1/2}(y_{i}-g(x_{i}))K_{x_{i}},(L_{K,t}+\lambda_t I)^{-1/2}(y_{i}-g(x_{i}))K_{x_{i}}\rangle_{K,t}\\
    &\leq
    4\tilde{M}^2\|(L_{K,t}+\lambda_t I)^{-1/2}(K_{{x_i}})\|_{K,t}^2
    = 4\tilde{M}^2\sum_{k=1}^\infty\frac{(\varphi_{k,t}({x_i}))^2}{\sigma_{k,t}+\lambda_t},
\end{align*}
where $(\sigma_{k,t},\varphi_{k,t})_{k=1}^\infty$ is the normalized eigen-pairs of $L_{K,t}$.
Thus, we have
$
     \|\eta_t(z_{i})\|_{K,t} \leq \frac{2\tilde{M}
     }
    {\sqrt{\lambda_t}}
$
and
\begin{align*}
 & E[ \|\eta_t(z_{i})\|_{K,t}^2]
   \leq
    4\tilde{M}^2E\left[\sum_{k=1}^\infty\frac{(\varphi_{k,t} {(x_i)}))^2}{\sigma_{k,t}+\lambda_t}\right]
    =
    4\tilde{M}^2 \sum_{k=1}^\infty\frac{E[\varphi_{k,t}^2 {(x_i)}]}{\sigma_{k,t}+\lambda_t}\\
    =&
    4\tilde{M}^2 \sum_{k=1}^\infty\frac{\sigma_{k,t}}{\sigma_{k,t}+\lambda_t}
    =4\tilde{M}^2\mbox{Tr}(L_{K,t}(L_{K,t}+\lambda_t I)^{-1})=
    4\tilde{M}^2\mathcal N_t(\lambda_t),
\end{align*}
where we used the fact $E[\varphi_{k,t}^2]=\sigma_{k,t}$ which was proven in \cite[eqs.(51)]{lin2017distributed}.
Therefore, it follows from Lemma \ref{Lemma:Pinelislemma}
that with confidence $1-\delta$, it holds that
$$
  \left\|\frac1{|D|}\sum_{i=1}^{|D|}\eta_t(z_{i})\right\|
  \leq {4\tilde{M} \log
 \bigl(2/{\delta}\bigr) \over {\sqrt{\lambda_t}} |D|}  + \sqrt{{ 8\tilde{M}^2\mathcal N_t(\lambda_t)
 \log \bigl(2/{\delta}\bigr) \over |D|}}.
$$
This completes the proof of Lemma \ref{Lemma:Concentration-1}. $\Box$

Setting  $y_i=y^*_{i,t}$ and $\tilde{M}=(T-t+1)M$ in Lemma \ref{Lemma:Concentration-1}, we can derive the following bound of $\mathcal P_{D,\lambda_t,t}$.
\begin{lemma}\label{Lemma:PD}
Let $0<\delta<1$. Under Assumption 1,
with confidence $1-\delta$, it holds that
$$
  \mathcal P_{D,\lambda_t,t}
  \leq
  \frac{ 2\sqrt{2}(T-t+1)M\log(2/\delta)}{\sqrt{|D|}}\left(
  \frac{\sqrt{2}}{  \sqrt{|D|\lambda_t}}+\sqrt{\mathcal N_t(\lambda_t)}\right).
$$
\end{lemma}

Unlike $ \mathcal P_{D,\lambda_t,t}$, the bounds of $\mathcal S_{D,\lambda_t,t}$ and $\overline{\mathcal S}_{D,\lambda_t,t}$  are much more sophisticated, which requires an upper bound of $\|Q_{D,\lambda_t,t}\|_{\mathcal L_t^\infty}$ and  $\|\overline{Q}_{D,\lambda_t,t}\|_{\mathcal L_t^\infty}$ for $\forall$ $t=1,\dots,T$. We leave them in the next subsection.

\subsection{Uniform bounds of Q-functions}
In this part, we aim at deriving $\|Q_{D,\lambda_t,t}\|_{\mathcal L_t^\infty}$ and  $\|\overline{Q}_{D,\lambda_t,t}\|_{\mathcal L_t^\infty}$  so that $y_{i,t}$ and $\overline{y}_{i,t}$ can also be uniformly bounded.
The following lemma presents an iterative relation between $\|Q_{D,\lambda_{t+1},t+1}\|_{\mathcal L_{t+1}^\infty}$ and
$\|Q_{D,\lambda_t,t}\|_{\mathcal L_t^\infty}$.

\begin{lemma}\label{Lemma:iteration:KRR}
Let  $0\leq \delta\leq 1$ satisfy
\begin{equation}\label{bound-delta}
    \delta\geq 8T\exp\left\{-\frac{2r+s} {4sC_1^*(\log (C_0+1)+2)} |D|^{\frac{2r+s-1}{8r+4s}}\log^{-1} |D|\right\}.
\end{equation}
Under Assumptions 1-4 with $\frac12\leq r\leq 1$ and $0<s\leq 1$, if $\lambda_t=|D|^{-\frac{1}{2r+s}}$ for $t=1,\dots,T$,  then with confidence $1-\delta/T$, it holds that
\begin{equation}\label{iterated-uniform-bound-KRR}
    \|Q_{D,\lambda_t,t}\|_{\mathcal L_t^\infty}+M\leq \bar{C}\sum_{\ell=t}^T(T-\ell+2)\left(2\mu^{1/2}\right)^{\ell-t}\left(\left\|Q_{D,\lambda_{\ell+1},\ell+1}\right\|_{\mathcal L_{\ell+1}^\infty}+M\right),\qquad t=1,2,\dots,T,
\end{equation}
where $\bar{C}$ is a constant depending only on $C_0$,  $\kappa$, $r$, $s$, and $\max_{t=1,\dots,T}\|h_t\|_{\mathcal L_t^2}$.
\end{lemma}

{\bf Proof.} Denote $\Phi_t=\|Q_{D,\lambda_t,t}\|_{\mathcal L_t^\infty}$.
Since $\lambda_1=\dots=\lambda_T=|D|^{-\frac1{2r+s}}$, we have from Assumption 4
that
$   \mathcal N_t(\lambda_t)\leq C_0\lambda_t^{-s}
   = C_0 |D|^{\frac{s}{2r+s}}$ for $\forall$ $t=1,2,\dots,T$.
Then
\begin{equation}\label{bound-media0111}
 \frac{1}{\sqrt{|D|}}\left( \frac{1}{  \sqrt{|D|\lambda_t}}+\sqrt{\mathcal N_t(\lambda_t)}\right)
 \leq (\sqrt{C_0}+1)|D|^{\frac{-r}{2r+s}},\qquad \forall t=1,2,\dots,T,
\end{equation}
and
\begin{equation}\label{Bound BD}
      \mathcal B_{D,\lambda_t,t}
      \leq
     \frac{2s}{2r+s} (\log (C_0+1)+2) |D|^{\frac{1-2r-s}{4r+2s}}\log |D|.
\end{equation}
Therefore,   Lemma \ref{Lemma:AD}  and Lemma \ref{Lemma:PD} yield that, with confidence $1-\delta/(4T)$ with $\delta$ satisfying \eqref{bound-delta}, there hold for any $t=1,\dots, T,$
\begin{align}
         \mathcal A_{D,\lambda_t,t} &\leq  2, \label{Bound-AD} \\
           \mathcal P_{D,\lambda_t,t}
     &\leq
     4(T-t+1)M(\sqrt{C_0}+1)|D|^{\frac{-r}{2r+s}}\log\frac{8T}\delta. \label{Bound-PD}
\end{align}
We get from Assumption 1
that $|{y}_{i,t}|\leq M+\Phi_{t+1}$ almost surely. Hence, it follows from (16),
Lemma \ref{Lemma:UD} with $f=Q_t^*-Q^\star_{D,\lambda_t,t}$, and
Lemma \ref{Lemma:Concentration-1} with $y_i={y}_{i,t}-y^*_{i,t}$ and $\tilde{M}=(T-t+2)M+\Phi_{t+1}$
that
\begin{align}
   \mathcal U_{D,\lambda_t,t,Q_t^*-Q^\star_{D,\lambda_t,t}}
   &\leq
    2((T-t+2)M+\Phi_{t+1})(\kappa+1)(\sqrt{C_0}+1)|D|^{\frac{-r}{2r+s}}\log\frac{8T}{\delta},        \label{Bound-UDT} \\
    \mathcal S_{D,\lambda_t,t}
    &\leq
    4((T-t+2)M+\Phi_{t+1})(\sqrt{C_0}+1)|D|^{\frac{-r}{2r+s}}\log\frac{8T}\delta,  \label{Bound-SDT}
\end{align}
hold with confidence $1-\delta/(4T)$.
Plugging \eqref{Bound-AD}, \eqref{Bound-PD}, \eqref{Bound-UDT}, and \eqref{Bound-SDT} into Proposition \ref{Proposition:Oracle-KRR}, we obtain from \eqref{bound-delta} that
\begin{align*}
  &\|Q_{D,\lambda_t,t}-Q_t^*\|_{K,t}\leq
    \lambda_t^{-1/2}\|(L_{K,t}+\lambda_tI)^{1/2}(Q_{D,\lambda_t,t}-Q_t^*)\|_{K,t}\\
    \leq&
     \sum_{\ell=t}^T(T- \ell+2) \left(2\mu^{1/2}\right)^{\ell-t}\left(\|h_\ell\|_{\mathcal L^2_\ell}+(\sqrt{C_0}+1)(M+\Phi_{\ell+1})( 2\kappa+10)\right)|D|^{\frac{-2r+1}{4r+2s}}\log\frac{8T}\delta\\
     \leq&
       \frac{(2r+s)\sum_{\ell=t}^T(T-\ell+2) \left(2\mu^{1/2}\right)^{\ell-t}\left(\|h_\ell\|_{\mathcal L^2_\ell}+\left(\sqrt{C_0}+1\right)(M+\Phi_{\ell+1})\left( 2\kappa+10\right)\right)} {(4sC_1^*(\log (C_0+1)+2))}\\
       \leq&
       \bar{C}_1\sum_{\ell=t}^T(T-\ell+2)\left(2\mu^{1/2}\right)^{\ell-t}(\Phi_{\ell+1}+M)
\end{align*}
holds with confidence $1-\delta/T$, where $\delta$ satisfies \eqref{bound-delta}, and
$$
 \bar{C}_1:=\frac{4(2r+s) \left(\max_{\ell=1,\dots,T}\|h_\ell\|_{\mathcal L^2_\ell}+(\sqrt{C_0}+1)\left( 2\kappa+10\right)\right)} {4sC_1^*(\log (C_0+1)+2)}.
$$
Therefore, we have
\begin{align*}
    &\|Q_{D,\lambda_t,t}\|_{\mathcal L_t^\infty}+M
    \leq
    \kappa  \|Q_{D,\lambda_t,t}\|_{K,t}+M
    \leq
    \kappa\|Q_{D,\lambda_t,t}-Q_t^*\|_{K,t}+\kappa\|Q_t^*\|_{K,t}+M\\
    \leq&
    \bar{C}\sum_{\ell=t}^T(T-\ell+2)\left(2\mu^{1/2}\right)^{\ell-t}(\Phi_{\ell+1}+M),
\end{align*}
where
$
   \bar{C}:= \kappa\bar{C}_1+\kappa^{2r}\max_{t=1,\dots,T}\|h_t\|_{\mathcal L_t^2}+1.
$
This completes the proof of Lemma \ref{Lemma:iteration:KRR}. $\Box$

Based on the above lemma, we can derive an upper bound of $\|Q_{D,\lambda_t,t}\|_{\mathcal L_t^\infty}$.

\begin{proposition}\label{Prop:uniform-bound-of-KRR}
Let $0\leq \delta\leq 1$ with $\delta$ satisfying \eqref{bound-delta}. Under Assumptions 1-4 with $\frac12\leq r\leq 1$ and $0<s\leq 1$, if $\lambda_t=|D|^{-\frac{1}{2r+s}}$ for $t=1,\dots,T$, then with confidence $1-\delta$, it holds that
$$
   \|Q_{D,\lambda_t,t}\|_{\mathcal L_t^\infty}
   \leq
  2 \bar{C}^T\left(2\mu^{1/2}\right)^{T-t} M\prod_{\ell=t}^{T-1}\left((T-\ell+2)\left(2\mu^{1/2}\right)^{\ell-t}+1\right)-M,\qquad t=1,\dots,T.
$$
\end{proposition}

{\bf Proof.}   Since for any $\xi_t,\eta_t>0$,
$
    \xi_t\leq  \sum_{\ell=t}^T \eta_\ell\xi_{\ell+1}
$
implies
$
\xi_t\leq \prod_{\ell=t}^{T-1}(\eta_\ell+1)\eta_T\xi_{T+1}.
$
Set $\xi_t= \|Q_{D,\lambda_t,t}\|_{\mathcal L_t^\infty}+M$ and
 {$\eta_t=\bar{C}(T-\ell+2)\left(2\mu^{1/2}\right)^{\ell-t}$}.
We have from \eqref{iterated-uniform-bound-KRR} and $Q_{D,\lambda_{T+1},T+1}=0$ that
$$
   \|Q_{D,\lambda_t,t}\|_{\mathcal L_t^\infty}+M
   \leq
   \prod_{\ell=t}^{T-1}\left(\bar{C}(T-\ell+2)\left(2\mu^{1/2}\right)^{\ell-t}+1\right)2\bar{C}\left(2\mu^{1/2}\right)^{T-t} M.
$$
This completes the proof Proposition \ref{Prop:uniform-bound-of-KRR}. $\Box$

Similar to Lemma \ref{Lemma:iteration:KRR}, we present an iteration relation between $\left\|\overline{Q}_{D,\lambda_t,t}\right\|_{\mathcal L_t^\infty}$ and $\left\|\overline{Q}_{D,\lambda_{t+1},t+1}\right\|_{\mathcal L_{t+1}^\infty}$.

\begin{lemma}\label{Lemma:iteration:DKRR}
Let  $0\leq \delta\leq 1$ satisfy
\begin{equation}\label{bound-delta-1}
    \delta\geq 40Tm\exp\left\{-\frac{2r+s} {4sC_1^*(\log (C_0+1)+2)} |D|^{\frac{2r+s-1}{16r+8s}}m^{-1/2}\log^{-1} |D|\right\}.
\end{equation}
Under Assumptions 1-4 with $\frac12\leq r\leq 1$ and $0<s\leq 1$, if $\lambda_t=|D|^{-\frac{1}{2r+s}}$ for $t=1,\dots,T$, $|D_1|=\dots=|D_m|$, and $m$ satisfies (22),
then with confidence $1-\delta/T$, it holds that
\begin{equation}\label{iterated-uniform-bound-DKRR}
    \|{\overline{Q}}_{D,\lambda_t,t}\|_{\mathcal L_t^\infty}+M\leq \hat{C}\sum_{\ell=t}^T(T-t+2)\left(2\mu^{1/2}\right)^{\ell-t}(\Phi_{\ell+1}+M),\qquad t=1,2,\dots,T,
\end{equation}
where $\bar{C}$ is a constant depending only on $C_0$,  $\kappa$, $r$, $s$, and $\max_{t=1,\dots,T}\|h_t\|_{\mathcal L_t^2}$.
\end{lemma}

{\bf Proof.} Write {$\Psi_t=\|\overline{Q}_{D,\lambda_t,t}\|_{\mathcal L_t^\infty}$}. Since $|D_1|=\dots=|D_m|$ and $\lambda_1=\dots=\lambda_T=|D|^{-\frac1{2r+s}}$,
(22) yields
\begin{equation}\label{bound-media0111-DKRR}
 \frac{1}{\sqrt{|D_j|}}\left( \frac{1}{  \sqrt{|D_j|\lambda_t}}+\sqrt{\mathcal N_t(\lambda_t)}\right)
 \leq (\sqrt{C_0}+1)\sqrt{m}|D|^{\frac{-r}{2r+s}},\qquad \forall ~ t=1,2,\dots,T,
\end{equation}
and
\begin{equation}\label{Bound-BDj}
      \mathcal B_{D,\lambda_t,t}
      \leq
     \frac{2s}{2r+s} (\log (C_0+1)+2) \sqrt{m}|D|^{\frac{1-2r-s}{4r+2s}}\log |D|.
\end{equation}
From Lemma \ref{Lemma:WD}, Lemma \ref{Lemma:AD}, and Lemma \ref{Lemma:PD} with $D=D_j$, we obtain that for any $\delta$ satisfying \eqref{bound-delta-1}, with confidence $1-\delta/(10mT)$, it holds that
\begin{align}
    \mathcal W_{D_j,\lambda_t,t}
         &\leq
         C_1^*\frac{2s}{2r+s} (\log (C_0+1)+2) \sqrt{m}|D|^{\frac{1-2r-s}{4r+2s}}\log |D| \log\frac{40mT}\delta,\label{Bound-WDj}\\
         \mathcal A_{D_j,\lambda_t,t}
         &\leq
         2,   \label{Bound-ADj}\\
         \mathcal P_{D_j,\lambda_t,t}
         &\leq
          4(T-t+1)M(\sqrt{C_0}+1)\sqrt{m}|D|^{\frac{-r}{2r+s}}\log\frac{20mT}\delta. \label{Bound-PDj}
\end{align}
We get from Assumption 1 that
$|\overline{y}_{i,t}|\leq M+\Psi_{t+1}$ almost surely. Hence, it follows from
(16), Lemma \ref{Lemma:UD} with $f=Q_t^*-\overline{Q}^\star_{D,\lambda_t,t}$, and
Lemma \ref{Lemma:Concentration-1} with $y_i=\overline{y}_{i,t}-y^*_{i,t}$ and $\tilde{M}=(T-t+2)M+\Psi_{t+1}$
that
\begin{align}
   \mathcal U_{D_j,\lambda_t,t,Q_t^*-\overline{Q}^\star_{D,\lambda_t,t}}
   &\leq
    2((T-t+2)M+\Psi_{t+1})(\kappa+1)(\sqrt{C_0}+1)\sqrt{m}|D|^{\frac{-r}{2r+s}}\log\frac{20mT}{\delta},        \label{Bound-UDj}\\
    \overline{\mathcal S}_{D_j,\lambda_t,t}
    &\leq
    4((T-t+2)M+\Psi_{t+1})(\sqrt{C_0}+1)\sqrt{m}|D|^{\frac{-r}{2r+s}}\log\frac{20mT}\delta  \label{Bound-SDj}
\end{align}
hold with confidence $1-\delta/(10mT)$. Hence, with confidence $1-\delta/(2mT)$ with $\delta$ satisfying \eqref{bound-delta-1}, we have from (22) that
\begin{align*}
   &\left(\sum_{j=1}^m\frac{|D_j|}{|D|}\mathcal W_{D_j,\lambda_t,t}  \mathcal A_{D_j,\lambda_t,t}
    +
    \mathcal A_{D,\lambda_t,t}\right)^{\ell-t}\\
    \leq&
    \left( C_1^*\frac{2s}{2r+s} (\log (C_0+1)+2) \sqrt{m}|D|^{\frac{1-2r-s}{2r+s}}\log |D| \log\frac{40mT}\delta
    +1\right)^{\ell-t}
    \leq
    2^{\ell-t}
\end{align*}
and
\begin{align*}
   &
   \lambda_{\ell}^{-1/2}\sum_{j=1}^m\frac{|D_j|}{|D|}\mathcal W_{D_j,\lambda_\ell,\ell}
      \mathcal A_{D_j,\lambda_\ell,\ell}  \left(\lambda_\ell^{r} \|h_\ell\|_{\mathcal L^2_\ell}+ \mathcal P_{D_j,\lambda_\ell,\ell}
   +
   \overline{\mathcal S}_{D_j,\lambda_\ell,\ell}
    +
    \mathcal U_{D_j,\lambda_\ell,\ell,Q^*_\ell-\overline{Q}^\star_{D_j,\lambda_\ell,\ell}}\right)\\
    \leq&
    \hat{C}_1 (T\!-\!\ell\!+\!2)(M\!+\!\Psi_{\ell+1})m|D|^{\frac{1-2r-s}{4r+2s}}\!\log |D|  |D|^{-\frac{2r-1}{4r+2s}} \log^2\frac{40mT}\delta\leq
   \!4\hat{C}_1(T\!-\!\ell\!+\!2)(M\!+\!\Psi_{\ell+1}),
\end{align*}
where $
   \hat{C}_1:=  \max_{t=1,\dots,T}\|h_t\|_{\mathcal L_t^2}+4(\sqrt{C_0}+1)+ 4(\sqrt{C_0}+1)
   +2(\kappa+1)(\sqrt{C_0}+1)
$.
Similarly, we can derive from \eqref{Bound-AD}, \eqref{Bound-PD}, \eqref{Bound-UDT}, and \eqref{Bound-SDT}  that, with confidence $1-\delta/(2T)$ with $\delta$ satisfying \eqref{bound-delta-1}, it holds that
\begin{align*}
    & \lambda_{\ell}^{-1/2}\mathcal A_{D,\lambda_\ell,\ell} \left(\lambda_\ell^{r}  \|h_\ell\|_{\mathcal L^2_\ell}+ \mathcal P_{D,\lambda_\ell,\ell}+ \overline{\mathcal S}_{D,\lambda_\ell,\ell}
    +\mathcal U_{D,\lambda_\ell,\ell,Q^*_\ell-\overline{Q}^\star_{D,\lambda_\ell,\ell}}\right)
     +
    \mathcal U_{D,\lambda_\ell,\ell,Q_\ell^*}\\
    \leq &
    \hat{C}_2\left(2\mu^{1/2}\right)^{\ell-t}({\Psi}_{\ell+1}+M),
\end{align*}
where $\hat{C}_2:=\bar{C}_1+1$.
The above three estimates together with {Proposition \ref{Prop:DKRR-Oracle}} yield that, with confidence $1-\delta/T$, it holds that
\begin{align*}
   &\left\|\overline{Q}_{D,{\lambda}_t,t}-Q_t^*\right\|_{K,t} \leq \lambda_t^{-1/2} \left\|(L_{K,t}+\lambda_t I)^{1/2}\left(\overline{Q}_{D,{\lambda}_t,t}-Q_t^*\right)\right\|_{K,t}\\
    \leq&
    (\hat{C}_1+2\hat{C}_2)
     \sum_{\ell=t}^T(T-\ell+2)\left(2\mu^{1/2}\right)^{\ell-t}({\Psi}_{\ell+1}+M).
\end{align*}
Therefore, we have
\begin{align*}
    &\|\overline{Q}_{D,\lambda_t,t}\|_{\mathcal L_t^\infty}+M
    \leq
    \kappa  \|\overline{Q}_{D,\lambda_t,t}\|_{K,t}+M
    \leq
    \kappa\|\overline{Q}_{D,\lambda_t,t}-Q_t^*\|_{K,t}+\kappa\|Q_t^*\|_{K,t}+M\\
    \leq&
    \hat{C}\sum_{\ell=t}^T(T-{\ell}+2)\left(\hat{C}_1 {2}\mu^{1/2}\right)^{\ell-t}({\Psi}_{\ell+1}+M),
\end{align*}
where
$
   \hat{C}:= \kappa(\hat{C}_1+2\hat{C}_2)+\kappa^{2r}\max_{t=1,\dots,T}\|h_t\|_{\mathcal L_t^2}+1.
$
This completes the proof of Lemma \ref{Lemma:iteration:DKRR}. $\Box$

Using the same approach as that in proving Proposition \ref{Prop:uniform-bound-of-KRR}, we can derive the following proposition.

\begin{proposition}\label{Prop:uniform-bound-of-DKRR}
Let $0\leq \delta\leq 1$ with $\delta$ satisfying \eqref{bound-delta-1}. Under Assumptions 1-4 with $\frac12\leq r\leq 1$ and $0<s\leq 1$, if $\lambda_t=|D|^{-\frac{1}{2r+s}}$ for $t=1,\dots,T$, $|D_1|=\dots=|D_m|$, and $m$ satisfies (22),
then with confidence $1-\delta$, it holds that
$$
   \|{\overline{Q}}_{D,\lambda_t,t}\|_{\mathcal L_t^\infty}
   \leq
  2 \hat{C}^T\left(2\mu^{1/2}\right)^{T-t} M\prod_{\ell=t}^{T-1}\left((T-\ell+2)\left(2\mu^{1/2}\right)^{\ell-t}+1\right)-M,\qquad t=1,\dots,T.
$$
\end{proposition}

 \subsection{Deriving generalization errors}

In this subsection, we derive the generalization error of KRR-DTR and DKRR-DTR. To prove our main theorems, we need the following lemma, which is standard in statistical learning theory.

\begin{lemma}\label{Lemma:prob-to-exp}
Let $0<\delta<1$, and $\xi\in\mathbb R_+$ be a random variable. If $\xi\leq u\log^b\frac{c}{\delta}$ holds with confidence $1-\delta$  for some $u,b,c>0$, then
$E[\xi]\leq c\Gamma(b+1) u$,
where $\Gamma(\cdot)$ is the Gamma function.
\end{lemma}

{\bf Proof.}
Since $\xi\leq u\log^b\frac{c}{\delta}$ holds with confidence $1-\delta$, we have
$ P[\xi>t]\leq c\exp\{-u^{-1/b}t^{1/b}\} $. Using the probability to expectation formula
\begin{equation}\label{expectation formula}
    E[\xi] =\int_0^\infty P\left[\xi > \varepsilon\right] d \varepsilon
\end{equation}
  to the positive random variable $\xi$, we have
$E[\xi]\leq  c\int_{0}^\infty\exp\{-u^{-1/b}\varepsilon^{1/b}\}d\varepsilon
     \leq cu\Gamma(b+1)$.
This completes the proof of Lemma \ref{Lemma:prob-to-exp}. $\Box$




With the above foundations, we are in a position to prove our main theorems.

{\bf Proof of Theorem 1.
}
It follows from Proposition \ref{Proposition:Oracle-KRR} that
\begin{align*}
     &E\left[\|(Q_{D,\lambda_t,t}-Q_t^*)\|_{\mathcal L_t^2}\right]
     \leq E\left[\|(L_{K,t}+\lambda_t I)^{1/2}(Q_{D,\lambda_t,t}-Q_t^*)\|_{K,t}\right] \\
    \leq&
    \sum_{\ell=t}^T \mu^\frac{\ell-t}2  E\left[ \mathcal A_{D,\lambda_\ell,\ell}^{\ell-t+1}   \left(\lambda_\ell^{r}\|h_\ell\|_{\mathcal L^2_\ell}
   +\mathcal P_{D,\lambda_\ell,\ell}
    +
      \mathcal S_{D,\lambda_\ell,\ell}+ \mathcal U_{D,\lambda_\ell,\ell,Q^*_\ell-Q^\star_{D,\lambda_\ell,\ell}} \right)\right].
\end{align*}
For $\lambda_1=\dots=\lambda_T=|D|^{-\frac{1}{2r+s}}$, it follows from \eqref{Bound-AD}, \eqref{Bound-PD},  \eqref{Bound-UDT}, \eqref{Bound-SDT}, and Proposition \ref{Prop:uniform-bound-of-KRR} that,
with confidence $1-\delta$ with $\delta$ satisfying \eqref{bound-delta}, it holds that
\begin{align*}
&
\mathcal A_{D,\lambda_\ell,\ell}^{\ell-t+1}   \left(\lambda_\ell^{r}\|h_\ell\|_{\mathcal L^2_t}
   +\mathcal P_{D,\lambda_\ell,\ell}
    +
      \mathcal S_{D,\lambda_\ell,\ell}+ \mathcal U_{D,\lambda_\ell,\ell,Q^*_\ell-Q^\star_{D,\lambda_\ell,\ell}}\right)\\
    \leq&
    2^{\ell-t}\tilde{C}_1\bar{C}^T\left(2\mu^{1/2}\right)^{T-\ell} M\prod_{k=\ell}^{T-1}\left((T-k+2)\left(2\mu^{1/2}\right)^{k-\ell}+1\right)
    |D|^{\frac{-r}{2r+s}
    }\log\frac{8T}{\delta},
\end{align*}
where
$   \tilde{C}_1:= 4\max_{t=1,\dots,T}\|h_t\|_{\mathcal L_t^2}+16M(\sqrt{C_0}+1)+2(\kappa+1)(\sqrt{C_0}+1)+4(\sqrt{C_0}+1)$.
Then it follows from Lemma \ref{Lemma:prob-to-exp}, \eqref{bound-delta},  and \eqref{expectation formula} that
\begin{align*}
    &E\left[\|Q_{D,\lambda_t,t}-Q_t^*\|_{\mathcal L_t^2}\right]
    \leq
    \int_{0}^\infty P\left[\left\|Q_{D,\lambda_t,t}-Q_t^*\right\|_{\mathcal L_t^2}>\varepsilon\right]d\varepsilon\\
    \leq&
    \int_{0}^{8T\exp\left\{-\frac{2r+s} {4sC_1^*(\log (C_0+1)+2)} |D|^{\frac{2r+s-1}{8r+4s}}\log^{-1} |D|\right\}} d\varepsilon\\
    &+
    \int_{8T\exp\left\{-\frac{2r+s} {4sC_1^*(\log (C_0+1)+2)} |D|^{\frac{2r+s-1}{8r+4s}}\log^{-1} |D|\right\}}^\infty   P\left[\|Q_{D,\lambda_t,t}-Q_t^*\|_{\mathcal L_t^2}>\varepsilon \right]d\varepsilon\\
    \leq&
     \tilde{C}_2T\sum_{\ell=t}^T \mu^\frac{\ell-t}2  2^{\ell-t}\bar{C}^T(2\mu^{1/2})^{T-\ell}  \prod_{k=\ell}^{T-1}\left((T-k+2)(2\mu^{1/2})^{k-\ell}+1\right)
    |D|^{\frac{-r}{2r+s}
    },
\end{align*}
where $\tilde{C}_2$ is a constant depending only on $\tilde{C}_1$, $M$, $r,s$, and $C_0$.
Noting further (18),
we then have
\begin{align*}
 & E\left[V_1^*(S_1)-V_{ {{\pi}_{D,\vec{\lambda}},1}}(S_1)\right] \\
  \leq &
  C_1  \sum_{t=1}^T \mu^{t/2} T\sum_{\ell=t}^T \mu^\frac{\ell-t}2  2^{\ell-t}\bar{C}^T\left(2\mu^{1/2}\right)^{T-\ell}  \prod_{k=\ell}^{T-1}\left((T-k+2)\left(2\mu^{1/2}\right)^{k-\ell}+1\right)|D|^{\frac{-r}{2r+s}
    }
\end{align*}
for $C_1=2\tilde{C}$. This completes the proof of Theorem 1.
$\Box$

{\bf Proof of Theorem 2.} It follows from Proposition \ref{Prop:DKRR-Oracle} that
\begin{align*}
    & E\left[\left\|\overline{Q}_{D,{\lambda}_t,t}-Q_t^*\right\|_{\mathcal L_t^2}\right] \leq E\left[\left\|(L_{K,t}+\lambda_t I)^{1/2}\left(\overline{Q}_{D,{\lambda}_t,t}-Q_t^*\right)\right\|_{K,t}\right]\\
    \leq&
     \sum_{\ell=t}^T\prod_{k=t+1}^\ell \left(\left(\sum_{j=1}^m\frac{|D_j|}{|D|}\mathcal W_{D_j,\lambda_k,k}  \mathcal A_{D_j,\lambda_k,k}
    +
    \mathcal A_{D,\lambda_k,k}\right) \mu^{1/2}\right)\\
    &\times
     \left(\sum_{j=1}^m\frac{|D_j|}{|D|}\mathcal W_{D_j,\lambda_\ell,\ell}
      \mathcal A_{D_j,\lambda_\ell,\ell}  \left(\lambda_\ell^{r} \|h_\ell\|_{\mathcal L^2_\ell}+ \mathcal P_{D_j,\lambda_\ell,\ell}
   +
   \overline{\mathcal S}_{D_j,\lambda_\ell,\ell}
    +
    \mathcal U_{D_j,\lambda_\ell,\ell,Q^*_\ell-\overline{Q}^\star_{D_j,\lambda_\ell,\ell}} \right)\right.\\
    &+
    \left.\left.\mathcal A_{D,\lambda_\ell,\ell}  \left(\lambda_\ell^{r}  \|h_\ell\|_{\mathcal L^2_\ell}+  \mathcal P_{D,\lambda_\ell,\ell}+   \overline{\mathcal S}_{D,\lambda_\ell,\ell}
    +\mathcal U_{D,\lambda_\ell,\ell,Q^*_\ell-\overline{Q}^\star_{D,\lambda_\ell,\ell}} \right)
     +
    \mathcal U_{D,\lambda_\ell,\ell,Q_\ell^*}\right)\right]
     .
\end{align*}
But  (22), \eqref{Bound-AD}, \eqref{bound-delta-1}, \eqref{Bound-ADj}, \eqref{Bound-PDj}, \eqref{Bound-UDj}, \eqref{Bound-SDj}, and Proposition \ref{Prop:uniform-bound-of-DKRR} yield that for any $t=1,\dots,T$, with confidence $1-\delta/2$, it holds that
$
   \left(\sum_{j=1}^m\frac{|D_j|}{|D|}\mathcal W_{D_j,\lambda_t,t}  \mathcal A_{D_j,\lambda_t,t}
    +
    \mathcal A_{D,\lambda_t,t}\right)
    \leq 2
$
and
\begin{align*}
   &\sum_{j=1}^m\frac{|D_j|}{|D|}\mathcal W_{D_j,\lambda_\ell,\ell}
      \mathcal A_{D_j,\lambda_\ell,\ell}  \left(\lambda_\ell^{r} \|h_\ell\|_{\mathcal L^2_\ell}+ \mathcal P_{D_j,\lambda_\ell,\ell}
   +
   \overline{\mathcal S}_{D_j,\lambda_\ell,\ell}
    +
    \mathcal U_{D_j,\lambda_\ell,\ell,Q^*_\ell-\overline{Q}^\star_{D_j,\lambda_\ell,\ell}} \right)\\
    \leq&
   \tilde{C}_3 m|D|^{\frac{1-2r-s}{4r+2s}}\log |D| |D|^{-\frac{r}{2r+s}}  \hat{C}^T \left(2\mu^{1/2}\right)^{T-t}  \prod_{\ell=t}^{T-1}\left((T-\ell+2)\left(2\mu^{1/2}\right)^{\ell-t}+1\right)\log\frac{40mT}  \delta\\
   \leq&
   \tilde{C}_3|D|^{-\frac{r}{2r+s}}  \hat{C}^T\left(2\mu^{1/2}\right)^{T-t}  \prod_{\ell=t}^{T-1}\left((T-\ell+2)\left(2\mu^{1/2}\right)^{\ell-t}+1\right)\log\frac{40}  \delta,
\end{align*}
where we use $\log ab\leq (\log a+1)\log b$ for $a,b\geq 3$ in the last inequality, and
$$
   \tilde{C}_3:=6MC_1^*\frac{2s}{2r+s} (\log (C_0+1)+2) \left(\max_{t=1,\dots,T}\|h_t\|_{\mathcal L_t^2}+8(\sqrt{C_0}+1)\right)
   +2(\kappa+1)(\sqrt{C_0}+1).
$$
 Similarly, we can derive that with confidence $1-\delta/2$, it holds that
 \begin{align*}
 &\mathcal A_{D,\lambda_\ell,\ell}  \left(\lambda_\ell^{r}  \|h_\ell\|_{\mathcal L^2_\ell}+  \mathcal P_{D,\lambda_\ell,\ell}+   \overline{\mathcal S}_{D,\lambda_\ell,\ell}
    +\mathcal U_{D,\lambda_\ell,\ell,Q^*_\ell-\overline{Q}^\star_{D,\lambda_\ell,\ell}} \right)
     +
    \mathcal U_{D,\lambda_\ell,\ell,Q_\ell^*}\\
    \leq&
     \tilde{C}_4|D|^{-\frac{r}{2r+s}}  \hat{C}^T \left(2\mu^{1/2}\right)^{T-t}  \prod_{\ell=t}^{T-1}\left((T-\ell+2)\left(2\mu^{1/2}\right)^{\ell-t}+1\right)\log\frac{40T}  \delta,
 \end{align*}
where
$\tilde{C}_4:=6M\left(\max_{t=1,\dots,T}\|h_t\|_{\mathcal
   L_t^2}+8(\sqrt{C_0}+1)\right)
   +2(\kappa+1)(\sqrt{C_0}+1)$.
Hence, we have from Lemma \ref{Lemma:prob-to-exp} that
\begin{align*}
   E\left[\left\|\overline{Q}_{D,{\lambda}_t,t}-Q_t^*\right\|_{\mathcal L_t^2}\right]
   \leq &
   40Tm\exp\left\{-\frac{2r+s} {4sC_1^*(\log (C_0+1)+2)} |D|^{\frac{2r+s-1}{16r+8s}}m^{-1/2}\log^{-1} |D|\right\}\\
   &+
   80(\tilde{C}_3\!+\!\tilde{C}_4)T|D|^{-\frac{r}{2r+s}} \hat{C}^T\left(2\mu^{1/2}\right)^{T-t}  \prod_{\ell=t}^{T-1}\left((T\!-\!\ell\!+\!2)\left(2\mu^{1/2}\right)^{\ell-t}\!+\!1\right)\\
   \leq&
   C_2T|D|^{-\frac{r}{2r+s}}  \hat{C}^T \left(2\mu^{1/2}\right)^{T-t}  \prod_{\ell=t}^{T-1}\left((T-\ell+2)\left(2\mu^{1/2}\right)^{\ell-t}+1\right),
\end{align*}
where $C_2$ is a constant depending only on $\max_{t=1,\dots,T}\|h_t\|_{\mathcal
   L_t^2}$, $C_0$, $\kappa$, $M$, $r$, and $s$.
This together with (18) completes the proof of Theorem 2.
$\Box$

\end{APPENDICES}


\bibliographystyle{informs2014} 
\bibliography{dql} 



\end{document}